\documentclass[11pt]{article}

\usepackage[preprint]{acl}

\usepackage{times}
\usepackage{latexsym}

\usepackage[T1]{fontenc}

\usepackage[utf8]{inputenc}

\usepackage{microtype}

\usepackage{inconsolata}
\usepackage{listings}
\usepackage{booktabs}
\usepackage{geometry}
\usepackage{hyperref}
\usepackage{graphicx}
\usepackage{amsmath, amssymb, amsfonts}
\usepackage{algorithm}
\usepackage{algorithmic}

\usepackage{amsthm}
\usepackage{tikz}
\usepackage{authblk}
\usepackage{xcolor}
\usetikzlibrary{shapes.geometric, arrows, arrows.meta, positioning, calc}
\usepackage{pifont}
\usepackage{placeins}
\usepackage{newunicodechar}
\newunicodechar{✗}{\ding{55}}
\newunicodechar{✓}{\ding{51}}

\usetikzlibrary{arrows.meta}
\usepackage{amssymb}

\usepackage[most]{tcolorbox}
\usepackage{multirow}

\newtcolorbox{promptbox}[1]{
    enhanced,
    breakable,
    colback=gray!10,      
    colframe=black,       
    coltitle=white,       
    fonttitle=\bfseries,  
    title=#1,
    width=\textwidth,     
    boxsep=5pt,
    left=5pt, right=5pt, top=5pt, bottom=5pt
}

\tikzstyle{policy} = [rectangle, rounded corners, minimum width=2.5cm, minimum height=1cm,text centered, draw=black, fill=blue!10, font=\small]
\tikzstyle{data} = [ellipse, minimum width=2cm, minimum height=0.8cm, text centered, draw=black, fill=green!10, font=\footnotesize]
\tikzstyle{env} = [rectangle, minimum width=2.5cm, minimum height=1cm, text centered, draw=black, fill=red!10, font=\small]
\tikzstyle{decision} = [diamond, minimum width=2cm, minimum height=1cm, text centered, draw=black, fill=yellow!10, font=\footnotesize]
\tikzstyle{process} = [rectangle, minimum width=2cm, minimum height=0.8cm, text centered, draw=black, fill=orange!10, font=\small]
\tikzstyle{arrow} = [thick,->,>=stealth]

\title{DeReAct: Decomposed Reasoning and Acting for Reliable AI Agents}

\author{
  Ajay Vohra$^1$ \quad
  Tao Chen$^2$\thanks{Corresponding author} \quad
  Neeti Narayan$^1$ \quad
  Caron Zhang$^2$ \\
  $^1$Amazon \quad
  $^2$Apple \\
  \texttt{ajayvohr@amazon.com, tao\_chen4@apple.com} \\
  \texttt{nneeti@amazon.com, tianlu\_zhang@apple.com}
}

\begin{document}
\maketitle

\begin{abstract}


ReAct-based agents typically rely on a single LLM policy to propose actions,
interact with the environment, and decide when a task is complete. This
coupling makes action authorization and completion control difficult to enforce
independently, allowing errors to propagate and unsupported completion claims
to terminate execution. We introduce DeReAct, a modular agent architecture
that externalizes two gating policies: a Critic that validates proposed actions
before execution, and a Context Manager that reconstructs an
environment-supported \textsc{State} and certifies task completion.

Across GAIA and SWE-bench Verified, DeReAct improves Pass@1 most for weaker
Brain models, with gains of 6.5--7.0 points for Qwen3-Coder-480B and
4.2--5.2 points for Claude Sonnet~4.5; gains diminish as Brain capability
increases. Trajectory and ablation analyses show that external gating is
effective when targeted failures are sufficiently prevalent and the
gating policy is itself sufficient. With Claude Opus~4.5, Pass@1 remains
comparable to ReAct, while DeReAct produces more evidence-complete and
constraint-satisfying trajectories, indicating that completion control can
trade earlier termination for stronger grounding. Overall, DeReAct improves 
weaker agents while retaining grounding benefits as models strengthen.

\end{abstract}

\section{Introduction}

ReAct~\citep{yao2023react} and related single-policy agents
~\citep{shinn2023reflexion,yao2023tree,chen2025goalact} commonly rely on one
policy to interleave reasoning, action, and observation throughout execution.
While simple and effective, this design couples task solving with two
reliability-critical decisions: whether a proposed action should reach the
environment and whether the task is complete. The task-performing policy
therefore authorizes its own actions and certifies its own completion.
Errors can consequently propagate through the trajectory, distort later
decisions, and trigger premature termination before sufficient evidence is
gathered or task requirements are satisfied, a recurring failure pattern in
LLM agent systems~\citep{cemri2025why}. This pattern is confirmed by our trajectory analyses, which show that some early completions are weakly grounded or explicitly guessed.

Existing methods address stages of this failure pattern. Runtime guardrails
and reviewer policies constrain actions before execution
~\citep{wang2025agentspec,ta-etal-2026-reinforced}, while self-correction and
verifier methods critique outputs or track task progress
~\citep{gou2024critic,chen2025smurfs,han2026vlaagui}. Context-management
systems compress or externalize long trajectories
~\citep{packer2023memgpt,kang2025acon}. However, these approaches do not jointly
separate action authorization and completion control from ReAct or ground both
in independently reconstructed, environment-supported evidence. Consequently,
invalid actions may reach the environment, while completion may rely on
unverified reasoning or prior errors.

To address this gap, we introduce DeReAct, an inference-time architecture that
decomposes the ReAct loop into a task-solving \textbf{Brain} and two
independently configurable gating policies: a \textbf{Critic} and a
\textbf{Context Manager} (CM). The Critic is a hard pre-execution gate that
blocks actions violating configured validity or task constraints. After each
executed action, the CM reconstructs a compact \textsc{State} from
environment-supported evidence, excluding the Brain's unverified reasoning,
and accepts completion only when the \textsc{State} supports it. DeReAct thus
separates action authorization and completion control from task reasoning.

We evaluate DeReAct on GAIA~\citep{mialon2023gaia} and SWE-bench
Verified~\citep{jimenez2024swebench} across three Brain capability levels.
DeReAct improves Pass@1 most for weaker Brains, with gains of 6.5--7.0 points
for Qwen3-Coder-480B and 4.2--5.2 points for Claude Sonnet 4.5. These gains
diminish with Brain capability; with Claude Opus 4.5, Pass@1 remains comparable
to ReAct. Pass@3 shows a complementary pattern: repeated-attempt behavior varies by
benchmark and capability, with stronger Brains tending toward broader success.

Ablations show that the Critic and CM contribute differently
across Brains and domains, and that intervention frequency alone does not
predict effectiveness. External gating is effective when the targeted failures are sufficiently prevalent and the gating policy is itself \emph{sufficient}: its prompt, model, and available context must support reliable intervention. Prompt-, model-, and context-level limitations can be diagnosed from trajectories.

To summarize, our contributions are threefold. We introduce DeReAct, which externalizes pre-execution action authorization and task-completion control around an environment-supported STATE. We evaluate it across benchmarks and Brain capability levels, analyzing task performance, trajectory grounding, and the roles of its gating policies. Finally, we introduce policy sufficiency to diagnose when external gating is effective across Brains and domains. 
\section{Related Work}
\label{sec:related_work}

\paragraph{Model--Scaffold Interactions.}
Recent work shows that agent performance depends on both base-model
capability and scaffolding. \citet{riosgarcia2026corral} find that the
former explains most performance variance, while scaffolding primarily
helps procedural reasoning. \citet{hao2026recreate} show that different
scaffolds solve partially disjoint SWE-bench problems and benefit from
stronger models during scaffold editing. Both treat the scaffold as an 
aggregate choice; DeReAct instead decomposes it into control policies and 
studies how their contributions vary with capability.

\paragraph{Pre-Execution Validation in Agent Loops.}
Prior work constrains agent behavior through runtime validation and
monitoring, including executable guardrails, formal verifiers, and
pre-execution reviewers~\citep{wang2025agentspec,xiang2025guardagent,
miculicich2025veriguard,bhat2026interwhen}.
Closest to DeReAct's Critic is Reinforced Agent~\citep{ta-etal-2026-reinforced}, which reviews proposed
tool calls before execution. DeReAct instead evaluates each intent--command
pair against the verified \textsc{State} and enforces rejection as a hard
gate, leaving the environment unchanged.

\paragraph{Completion Verification and Termination Control.}
Premature termination and inadequate task verification are recurring failure
modes in agent systems~\citep{cemri2025why}. Prior work explores
self-correction with model or tool feedback~\citep{madaan2023selfrefine,
gou2024critic}, though such correction is unreliable without external
feedback~\citep{huang2024cannot}. Other systems introduce explicit completion
verification~\citep{chen2025smurfs,han2026vlaagui}.
DeReAct applies completion control to the full trajectory of a general ReAct
agent: the Context Manager reconstructs an environment-supported \textsc{State}
that excludes the Brain's unverified reasoning and ratifies a final answer
only when that state supports task completion.

\paragraph{Context Management and Agent Memory.}
Long agent trajectories motivate context selection, compression, and
externalization~\citep{liu2024lost}. MemGPT~\citep{packer2023memgpt}
manages in-context and archival memory, while ACON~\citep{kang2025acon}
compresses trajectory history. DeReAct's \textsc{State} is neither a general-purpose summary nor
a memory store: it is rebuilt from environment-supported evidence
at each step and paired with a separate sliding window of recent
raw history.

\paragraph{Multi-Agent and Loop Decomposition.}
Multi-agent frameworks such as AutoGen~\citep{wu2023autogen} and
MetaGPT~\citep{hong2024metagpt} distribute sub-tasks across model instances,
whereas DeReAct decomposes the control loop of a single agent.
Smurfs~\citep{chen2025smurfs} is closest in spirit, but decomposes branching
search and verifies sub-tasks, while DeReAct separates reasoning,
action authorization, verified-state construction, and completion control
within a linear ReAct loop.

\section{The DeReAct Architecture}
\label{sec:architecture}

Standard ReAct~\citep{yao2023react} uses a single policy that, at each step,
produces a reasoning trace (\textit{Thought}) and an action (\textit{Action});
the resulting observation is appended to the trajectory until the model emits
a \textit{Final Answer}. Thus, the same policy both selects actions and
declares task completion, against a history that interleaves environment
feedback with its own reasoning.

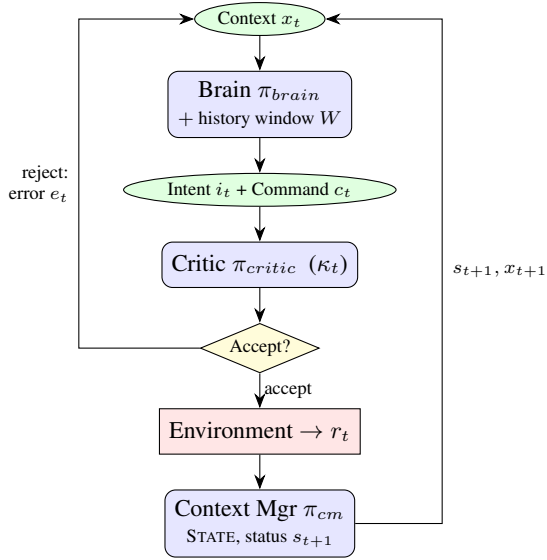
\begin{figure}[t]
\centering
\begin{tikzpicture}[font=\footnotesize, node distance=4.6mm, >={Stealth[length=2mm]},
  policy/.style={rectangle, rounded corners, draw, fill=blue!10, minimum width=2.2cm, minimum height=0.6cm, align=center},
  data/.style={ellipse, draw, fill=green!12, align=center, inner sep=1.5pt, font=\scriptsize},
  envv/.style={rectangle, draw, fill=red!10, minimum width=2.2cm, minimum height=0.6cm, align=center},
  dec/.style={diamond, draw, fill=yellow!18, aspect=2.4, inner sep=1pt, align=center, font=\scriptsize}]
\node[data] (ctx) {Context $x_t$};
\node[policy, below=of ctx] (brain) {Brain $\pi_{brain}$\\ \scriptsize $+$ history window $W$};
\node[data, below=of brain] (prop) {Intent $i_t$ + Command $c_t$};
\node[policy, below=of prop] (critic) {Critic $\pi_{critic}$\ \ ($\kappa_t$)};
\node[dec, below=of critic] (acc) {Accept?};
\node[envv, below=of acc] (env) {Environment $\to r_t$};
\node[policy, below=of env] (cm) {Context Mgr $\pi_{cm}$\\ \scriptsize \textsc{State}, status $s_{t+1}$};
\draw[->] (ctx)--(brain);
\draw[->] (brain)--(prop);
\draw[->] (prop)--(critic);
\draw[->] (critic)--(acc);
\draw[->] (acc)-- node[right,font=\scriptsize,inner sep=1.5pt]{accept} (env);
\draw[->] (env)--(cm);
\draw[->] (acc.west) -- ++(-1.60,0) |- node[pos=0.25,left,font=\scriptsize,align=right]{reject:\\error $e_t$} (ctx.west);
\draw[->] (cm.east) -- ++(1.15,0) |- node[pos=0.25,right,font=\scriptsize]{$s_{t+1},x_{t+1}$} (ctx.east);
\end{tikzpicture}
\caption{DeReAct control flow. The Brain proposes an
intent and command; the Critic gates execution; and
the Context Manager reconstructs the verified
\textsc{State} and determines task completion. Only
the Brain additionally receives a bounded history
window.}
\label{fig:dereact}
\end{figure}

DeReAct decomposes this loop into three independently prompted modules
(Figure~\ref{fig:dereact}): a \emph{Brain}, a \emph{Critic}, and a \emph{Context Manager} (CM). The Brain proposes an intent and executable
command. The Critic validates the command against configured validity and task constraints before execution, not its strategic correctness; rejected actions leave the environment unchanged and return feedback to the Brain. After each
accepted action, the CM reconstructs a verified
\textsc{State} from environment feedback and sets the task status to
\textsc{Continue} or \textsc{Success}. The \textsc{State} excludes the Brain's
unverified reasoning and is shared by all modules. Algorithm~\ref{alg:dereact} summarizes the execution loop.

\begin{algorithm}[t]
\caption{DeReAct Execution (agent-controlled loop)}
\label{alg:dereact}
\small
\begin{algorithmic}[1]
\STATE \textbf{Input:} Task $\mathcal{T}$, models $(\pi_{brain},\pi_{critic},\pi_{cm})$, Critic confidence threshold $\kappa_{\min}$; step limit $T_{\max}$; history-window size $W$ and initial-step count $m$
\STATE \textbf{Init:} $t\!\leftarrow\!0$, $x_0\!\leftarrow\!\text{InitialContext}(\mathcal{T})$, $s_0\!\leftarrow\!\textsc{Continue}$
\WHILE{$s_t=\textsc{Continue}$ \textbf{and} $t<T_{\max}$}
    \STATE $(i_t,c_t)\sim\pi_{brain}(\cdot\mid x_t,h_t)$ \quad\COMMENT{$h_t$: first $m$ $+$ most recent $W{-}m$ steps}
    \STATE $(d_t,\kappa_t)\sim\pi_{critic}(\cdot\mid x_t,i_t,c_t)$
    \IF{$d_t=0$ \textbf{or} $\kappa_t<\kappa_{\min}$}
        \STATE $e_t\sim\pi_{critic}(\text{error}\mid x_t,i_t,c_t)$
        \STATE $x_{t+1}\leftarrow(x_t,e_t)$;\ \ $s_{t+1}\leftarrow\textsc{Continue}$ \quad\COMMENT{no environment change}
    \ELSE
        \STATE $r_t\leftarrow\text{Env.execute}(c_t)$
        \STATE $(s_{t+1},x_{t+1})\sim\pi_{cm}(\cdot\mid x_t,i_t,c_t,r_t)$
    \ENDIF
    \STATE $t\leftarrow t+1$
\ENDWHILE
\end{algorithmic}
\end{algorithm}



\paragraph{Brain ($\pi_{brain}: (\mathcal{X}\times\mathcal{H})\!\rightarrow\!\Delta(\mathcal{I}\times\mathcal{C})$).}
Given context $x_t$ and a sliding window $h_t$ of raw execution history, the Brain produces a strategic intent $i_t$ and executable command $c_t$. To reduce long-context degradation~\citep{liu2024lost}, $h_t$ retains the first $m$ steps and the most recent $W{-}m$ steps, trimming the middle. Only the Brain receives this raw-history window; the Critic and CM operate without it. The Brain uses the same Thought--Action prompting as ReAct so performance differences reflect the surrounding architecture rather than prompt design.



\paragraph{Critic ($\pi_{critic}: (\mathcal{X}\times\mathcal{I}\times\mathcal{C})\!\rightarrow\!\Delta(\mathcal{D}\times\mathcal{K})$).}
Given context $x_t$, intent $i_t$, and proposed command $c_t$, the Critic
returns an accept/reject decision $d_t$ and confidence $\kappa_t$.
The command executes only if $d_t=1$ and $\kappa_t\geq\kappa_{\min}$;
otherwise, it is rejected and feedback $e_t$ is returned to the Brain.
DeReAct therefore implements a fail-closed pre-execution gate.
The Critic checks configured pre-execution constraints such as safety, syntax,
and explicit task requirements, but not strategy or task progress, providing
an independent validation layer that may use a smaller model than the Brain.

\textbf{Context Manager}
($\pi_{\mathrm{cm}}:(\mathcal{X}\times\mathcal{I}\times\mathcal{C}\times\mathcal{R})
\rightarrow\Delta(\mathcal{S}\times\mathcal{X})$).
After command $c_t$ returns result $r_t$, the CM reconstructs the verified
context $x_{t+1}$ (\textsc{State}) and sets
$s_{t+1}\in\{\textsc{Success},\textsc{Continue}\}$. The \textsc{State}
contains only environment-supported facts, excluding the Brain's unverified
reasoning, and is replaced rather than appended each step to avoid reverting
to the interleaved trajectory it replaces.

The raw-history window is available only to the Brain, not to the Critic or CM. This separates verified evidence from recent procedural detail. 
After each executed action, the CM returns \textsc{Success} only if the \textsc{State} contains sufficient evidence that the task-completion criteria are satisfied; otherwise, it returns \textsc{Continue}. The execution loop terminates only on \textsc{Success}.

\section{Experiments}
\label{sec:experiments}

We evaluate DeReAct against ReAct and targeted ablations on GAIA and
SWE-bench Verified across three Brain capability levels, measuring task
performance, control-policy effectiveness, trajectory grounding, and inference
efficiency.

\subsection{Setup}
\label{sec:setup}


We evaluate on two benchmarks. \textbf{GAIA}~\citep{mialon2023gaia}
contains 165 retrieval-and-reasoning validation questions scored by quasi-exact
match. \textbf{SWE-bench Verified}~\citep{jimenez2024swebench} evaluates
real-world GitHub issue resolution; we use the 489 instances supported by the
official test harness (Appendix~\ref{app:excluded}). Together, the benchmarks cover retrieval- and tool-assisted reasoning and execution-verifiable coding. Task success is measured with benchmark-native, non-LLM scoring; the judge in Section~\ref{subsec:grounded_exploration} assesses trajectory grounding only.


We run three attempts per task and report \textbf{Pass@1}, the mean
single-attempt success rate, and \textbf{Pass@3}, the fraction solved in at
least one attempt. \textbf{Lift} denotes DeReAct's improvement over ReAct,
and \textbf{Diff} the Pass@3 lift minus the Pass@1 lift. We compare
\textbf{ReAct} (Brain only), \textbf{DeReAct} (Brain + Critic + CM),
\textbf{Ablation A} (Brain + CM), and \textbf{Ablation B} (Brain + Critic).
With a CM, the Brain receives the current \textsc{State}, the first $m=5$
steps, and the most recent $W-m$ steps; otherwise the full raw trajectory is
used.

We evaluate Qwen3-Coder-480B~\citep{qwen3}, Claude Sonnet~4.5,
and Claude Opus~4.5, spanning increasing Brain capability.
The Critic uses Claude Haiku~4.5 throughout. Main Qwen3 and Sonnet
runs use a Haiku CM with $W=25$, while Opus uses an Opus CM with
$W=50$; Section~\ref{subsec:sufficiency} analyzes CM, prompt, and
window variants. We set $\kappa_{\min}=0.7$, $T_{\max}=1000$, and
temperature $0.1$. Prompts are benchmark-specific and fixed within
each benchmark (Appendix~\ref{app:prompts}).

We report standard errors across the three runs and assess task-level
uncertainty with two-sided paired permutation tests over tasks
($10{,}000$ iterations), keeping each task's three attempts together.
Tests are applied to Pass@1, Pass@3, and Diff; reported $p$-values are
uncorrected.

\begin{table}[tbp]
\centering\small
\setlength{\tabcolsep}{2pt}
\caption{
Main DeReAct performance across Brain capability levels. 
Full results:
Table~\ref{tab:gaia}, \ref{tab:gaia-pass3},
\ref{tab:swe_improved}, and~\ref{tab:swe_improved-pass3}.
}
\label{tab:compression}

\begin{tabular}{@{}lcccccc@{}}
\toprule
& \multicolumn{3}{c}{\textbf{Pass@1 (\%)}}
& \multicolumn{3}{c}{\textbf{Pass@3 (\%)}} \\
\cmidrule(lr){2-4}\cmidrule(l){5-7}
\textbf{Brain}
& \textbf{DeReAct} & \textbf{lift} & \textbf{$p$}
& \textbf{DeReAct} & \textbf{lift} & \textbf{$p$} \\ \hline

\multicolumn{7}{@{}l}{\emph{GAIA}} \\
Qwen3  & $41.4\pm3.3$ & $+6.5$ & .010 & $55.2\pm3.9$ & $+7.3$ & .068 \\
Sonnet & $54.5\pm3.4$ & $+4.2$ & .170 & $68.5\pm3.6$ & $+7.3$ & .060 \\
Opus   & $62.6\pm3.4$ & $+1.0$ & .697 & $73.3\pm3.4$ & $+2.4$ & .506 \\

\multicolumn{7}{@{}l}{\emph{SWE-bench Verified}} \\
Qwen3  & $44.0\pm1.8$ & $+7.0$ & ${<}.001$ & $58.7\pm2.2$ & $+4.5$ & .014 \\
Sonnet & $54.1\pm0.4$ & $+5.2$ & ${<}.001$ & $66.3\pm2.1$ & $+3.5$ & .039 \\
Opus   & $65.6\pm0.8$ & $-0.3$ & .753 & $71.6\pm2.0$ & $+1.4$ & .269 \\
\bottomrule
\end{tabular}
\end{table}

\subsection{Performance by Brain Capability}
\label{subsec:brain_capability}

Table~\ref{tab:compression} presents the main results. DeReAct significantly
improves Pass@1 for the Qwen3 and Sonnet Brains on both benchmarks, except for
Sonnet on GAIA, where the gain is positive but not significant. Under the Opus
Brain, Pass@1 is within noise of ReAct on both. Pass@1 gains therefore appear
to narrow as Brain capability increases. The same trend holds when all three Brains use a fixed Haiku CM and $W=25$ (Table ~\ref{tab:gaia} and~\ref{tab:swe_improved}).

Pass@3 shows a more nuanced pattern. On SWE-bench, Qwen3 and Sonnet have
larger Pass@1 than Pass@3 gains, suggesting improved consistency on already
solvable instances; on GAIA, their Pass@3 gains are larger. Under Opus,
Pass@3 gains exceed Pass@1 gains on both benchmarks, indicating broader
success across attempts despite unchanged single-attempt performance.
Sections~\ref{subsec:sufficiency} and~\ref{subsec:grounded_exploration}
examine this further.

\subsection{Contributions of the Critic and CM}
\label{subsec:critic_cm_contribution}

\begin{table}[t]
\centering
\small
\setlength{\tabcolsep}{2pt}
\caption{Critic/CM activation rates and Pass@1 lift by module. 
Full results: Table~\ref{tab:gaia_rej_full}
and~\ref{tab:swe_rej_full}.}
\label{tab:module_contributions}

\begin{tabular}{l rr rr rrr}
\toprule
& \multicolumn{2}{c}{\textbf{Critic act. (\%)}}
& \multicolumn{2}{c}{\textbf{CM act. (\%)}}
& \multicolumn{3}{c}{\textbf{Pass@1 $\Delta$ (pp)}} \\
\cmidrule(lr){2-3}\cmidrule(lr){4-5}\cmidrule(lr){6-8}
\textbf{Brain}
& \textbf{Abl.~B} & \textbf{Full}
& \textbf{Abl.~A} & \textbf{Full}
& \textbf{Abl.~B} & \textbf{Abl.~A} & \textbf{Full} \\
\midrule

\multicolumn{8}{l}{\textit{GAIA}} \\
Qwen3  & 4.3 & 28.1 & 92.4 & 93.3 & +0.7 & +4.9 & +6.5 \\
Sonnet & 0.2 & 2.2  & 89.2 & 90.4 & +1.2 & +4.4 & +4.2 \\
Opus   & 0.1 & 1.3  & 54.8 & 55.7 & $-2.2$ & +1.0 & +1.0 \\
\midrule

\multicolumn{8}{l}{\textit{SWE-bench Verified}} \\
Qwen3  & 32.6 & 16.2 & 93.1 & 93.1 & +6.6 & $-3.5$ & +7.0 \\
Sonnet & 1.7  & 2.1  & 65.3 & 72.4 & +2.4 & +3.3 & +5.2 \\
Opus   & 2.0  & 2.2  & 20.7 & 20.1 & +0.2 & $-0.4$ & $-0.3$ \\
\bottomrule
\end{tabular}
\end{table}


Table~\ref{tab:module_contributions} reports module activation rates and
ablation-based Pass@1 contributions. Critic activation is the fraction of
actions rejected, and CM activation the fraction of final answers vetoed.
Rates are shown for the corresponding ablations and full DeReAct, where both
policies use the verified \textsc{State}. Their individual lifts need not sum
to the joint lift because the policies interact.


The \textsc{State} changes what the Critic catches in a domain-dependent way:
for Qwen3, rejection rises from 4.3\% to 28.1\% on GAIA but falls from 32.6\%
to 16.2\% on SWE-bench, filtering spurious rejections. For Sonnet and Opus,
Critic activation remains near 2\%. Its accuracy gain also depends on overlap
with the CM: for Qwen3 it adds $+6.6$ Pass@1 points on SWE-bench but only
$+0.7$ on GAIA, where CM activation is already near-total. For Opus on GAIA,
the raw-history Critic hurts performance ($-2.2$), while full DeReAct
matches the CM-only lift.

The CM replaces the growing trajectory with a compact \textsc{State} and vetoes
unsupported completion. Its activation depends mainly on whether the Brain can
already ground completion, not on the Critic. On SWE-bench, Sonnet and Opus
test appropriately on 80.1\% and 78.2\% of trajectories
(Table~\ref{tab:swe_react_grounding}), and Opus CM vetoes only 20.1\% of final
answers versus 55.7\% on GAIA. Qwen3, by contrast, tests appropriately on only
4.6\% of SWE-bench trajectories and is vetoed at about 93\% on both benchmarks,
yet its CM lift is $+4.9$ on GAIA and $-3.5$ on SWE-bench. Thus, activation
alone does not explain accuracy gains.

\subsection{Policy Activation and Sufficiency}
\label{subsec:sufficiency}

\begin{table}[t]
\centering
\small
\setlength{\tabcolsep}{3pt}
\caption{Policy-sufficiency diagnostics for the Opus Brain.
$\star$ marks the main-results configuration.
\textsc{FAV} denotes Final Answer Validation.
For SWE-bench Verified, \textsubscript{min} denotes the one-line base completion
criterion and \textsubscript{evid} the evidence-aware improved prompt
(\ref{app:prompts:cm:swe}).}
\label{tab:policy_sufficiency}

\begin{tabular}{lccccc}
\toprule
\textbf{CM}
& \textbf{$W$}
& \textbf{Veto}
& \textbf{STATE}
& \textbf{FAV}
& \textbf{Pass@1} \\
& 
& \textbf{(\%)}
& \textbf{good (\%)}
& \textbf{poor (\%)}
& \textbf{lift (pp)} \\
\midrule

\multicolumn{6}{l}{\textit{GAIA}} \\
Haiku
& 25 & 81.7 & 80.2 & 19.8 & $-3.0$ \\
Opus
& 25 & 38.5 & 93.9 & 22.2 & $+0.2$ \\
Haiku
& 50 & 95.9 & 79.6 & 25.8 & $-5.0$ \\
Opus$^\star$
& 50 & 55.7 & 97.0 & 16.0 & $+1.0$ \\

\midrule

\multicolumn{6}{l}{\textit{SWE-bench Verified}} \\
Haiku$_{\mathrm{min}}$
& 25 & 10.2 & 83.0 & 8.0 & $-4.7$ \\
Haiku$_{\mathrm{evid}}$
& 25 & 56.5 & 88.0 & 2.0 & $-1.8$ \\
Haiku$_{\mathrm{evid}}$
& 50 & 59.3 & 83.0 & 3.0 & $-1.1$ \\
Opus$_{\mathrm{evid}}$
& 25 & 19.6 & 94.0 & 0.0 & $-2.4$ \\
Opus$_{\mathrm{evid}}^\star$
& 50 & 20.1 & 97.0 & 3.0 & $-0.3$ \\

\bottomrule
\end{tabular}
\end{table}

We use \emph{policy sufficiency} to describe how well a gating policy is
suited to a setting, treating it as continuous rather than binary. For the CM,
sufficiency depends on three adjustable levers: the completion prompt, which
controls veto behavior; the CM model, which determines \textsc{State}
fidelity; and the Brain's history window, which determines whether required
evidence remains available. We assess the first two with an independent
Claude Opus~4.7 judge, blind to benchmark outcomes, on 100 sampled complete
trajectories per configuration. \textsc{State Good} measures preservation of
verified evidence, while \textsc{FAV Poor} measures unsupported Final Answer
Validation. Together with veto rates and trajectory lengths, these diagnose prompt, model, and context limitations from trajectories.

High activation can still be harmful. On SWE-bench, the Haiku CM
vetoes 93.1\% of Qwen3 final answers yet lowers Pass@1 by 3.5 points, with
only 37\% \textsc{State Good} and 33\% \textsc{FAV Poor}
(Table~\ref{tab:cm_quality}). Under Opus
(Table~\ref{tab:policy_sufficiency}), improving the completion prompt cuts
\textsc{FAV Poor} from 8\% to 2\% and improves the lift from $-4.7$ to
$-1.8$; widening the history window raises it to $-1.1$, and scaling the CM brings it to $-0.3$ at 97\% \textsc{State Good}.

The levers are complementary rather than interchangeable. On SWE-bench,
scaling the CM before widening the window reduces the lift from $-1.8$ to
$-2.4$ at $W{=}25$: higher \textsc{State} fidelity cannot recover evidence
the Brain no longer retains. On GAIA, the CM model is instead the main
bottleneck, improving the lift from $-3.0$ to $+0.2$ at $W{=}25$ and from
$-5.0$ to $+1.0$ at $W{=}50$, while raising \textsc{State Good} from about
80\% to 94--97\%. \textsc{FAV Poor} nevertheless remains higher on GAIA
(16--26\% vs. 0--8\% on SWE-bench), suggesting completion is harder to
validate without an executable testing harness.

These diagnostics suggest a trajectory-based tuning strategy. If the Brain's
history window is short relative to the median ReAct trajectory length
(Tables~\ref{tab:gaia_cost},~\ref{tab:swe_cost}), increasing it can mitigate
context loss. Low \textsc{State Good} motivates scaling the CM model, with
${\sim}90\%$ serving as a useful reference point in our configurations
(Table~\ref{tab:cm_quality}). High \textsc{FAV Poor} motivates refining the
completion prompt and comparing the fraction of DeReAct trajectories with at
least one veto against ReAct trajectories that complete without grounding
evidence. \textsc{FAV Poor} below ${\sim}5\%$ was a useful target where
attainable; we reach 3\% on SWE-bench Verified but not on GAIA. These are
empirical diagnostics rather than universal thresholds.

\subsection{Grounded Exploration}
\label{subsec:grounded_exploration}

\begin{table}[t]
\caption{
Resolution after CM completion vetoes on matched
\emph{(instance, attempt)} pairs. R-only is ReAct resolution among DeReAct
failures; Grounded is the grounded fraction. Full results:
Tables~\ref{tab:gaia_veto_yield} and~\ref{tab:swe_veto_yield}.}
\label{tab:yield-excerpt}
\centering
\small
\setlength{\tabcolsep}{2.2pt}

\begin{tabular}{@{}lrrrrr@{}}
\toprule
\textbf{Brain}
& \textbf{DeReAct}
& \textbf{ReAct}
& \textbf{lift}
& \textbf{R-only}
& \textbf{Grounded} \\
\midrule

\multicolumn{6}{l}{\textit{GAIA}} \\
Qwen3      & 30.2 & 23.1 & $+7.1$ & 11.5 & 55 \\
Sonnet 4.5 & 43.1 & 35.6 & $+7.5$ & 23.5 & 44 \\
Opus 4.5   & 39.2 & 35.8 & $+3.4$ & 15.6 & 20 \\
\midrule

\multicolumn{6}{l}{\textit{SWE-bench Verified}} \\
Qwen3      & 43.7 & 37.3 & $+6.4$ & 17.9 & 9.2 \\
Sonnet 4.5 & 48.9 & 39.2 & $+9.7$ & 17.6 & 70.0 \\
Opus 4.5   & 59.3 & 61.0 & $-1.7$ & 17.8 & 61.9 \\

\bottomrule
\end{tabular}
\end{table}

We first ask whether CM vetoes promote productive exploration rather than
merely longer trajectories. Table~\ref{tab:yield-excerpt} compares DeReAct
and ReAct on matched \emph{(instance, attempt)} pairs where DeReAct has at
least one CM veto. DeReAct achieves higher resolution rates for Qwen3 and
Sonnet on both benchmarks ($+7.1$ and $+7.5$ points on GAIA; $+6.4$ and
$+9.7$ on SWE-bench Verified). Ablation~A shows the same pattern
(Tables~\ref{tab:gaia_veto_yield} and~\ref{tab:swe_veto_yield}), showing that it
persists without the Critic.

For Opus, the difference narrows to $+3.4$ points on GAIA and $-1.7$ on
SWE-bench. However, ReAct's additional resolutions are not uniformly
grounded: among pairs unresolved by DeReAct, ReAct resolves 15.6\% on GAIA
and 17.8\% on SWE-bench, of which only 20.0\% and 61.9\%, respectively, are
grounded in evidence. This indicates that some ReAct successes arise from early completion with weakly supported or guessed answers rather than fully grounded evidence.

We next examine whether DeReAct improves consistency or broadens success
across repeated attempts using
$\mathrm{Diff}=\Delta\mathrm{Pass@3}-\Delta\mathrm{Pass@1}$.
Negative \emph{Diff} indicates consistency gains, whereas positive values
indicate broader success across attempts. On SWE-bench, Qwen3 and Sonnet
have negative \emph{Diff} ($-2.5$ and $-1.7$), while Opus is positive across
all configurations on both benchmarks (Figure~\ref{fig:pass3}). Thus, with
stronger Brains, DeReAct shifts from consistency gains toward broader
exploration.

Finally, an independent Claude Opus~4.7 judge, blind to benchmark outcomes,
evaluates Opus~4.5 trajectories for evidence completeness, constraint
satisfaction, and admitted guessing on GAIA, and appropriate testing on
SWE-bench (Appendix~\ref{app:prompts:judge}). Tables~\ref{tab:gaia_grounding} and~\ref{tab:swe_grounding} further confirm this pattern. On GAIA, DeReAct increases evidence completeness and constraint satisfaction while reducing admitted guessing, with the largest
gains among incorrect trajectories. SWE-bench shows smaller but consistent
gains in appropriate testing, including on unresolved trajectories.

\subsection{Cost and Efficiency}
\label{subsec:cost}

DeReAct adds inference cost and can lengthen trajectories through the Critic and Context Manager. Cost and step distributions (Tables~\ref{tab:gaia_cost} and~\ref{tab:swe_cost}) show the same broad pattern: overhead is largest with weaker Brains and narrows as Brain capability increases, especially on SWE-bench Verified. On GAIA, mean per-run cost is about 2.9$\times$ ReAct with the Opus Brain and step counts retain a heavier tail; on SWE-bench, the strongest Opus configurations approach or match ReAct in mean cost despite the additional gating policies.

This domain difference likely reflects task structure: SWE-bench provides
executable test feedback as a natural completion signal, whereas GAIA permits
plausible answer submission without equally strong external verification.
Consistent with Section~\ref{subsec:grounded_exploration}, part of DeReAct's
overhead may therefore come from additional evidence gathering after
completions that ReAct would otherwise accept with weak grounding or explicit
guessing.

\begin{table}[t]
\centering\small
\setlength{\tabcolsep}{4pt}
\caption{Trajectory grounding on GAIA for Opus Brain.
``Both'' denotes evidence-complete \emph{and} constraints-satisfied.}
\label{tab:gaia_grounding}

\begin{tabular}{@{}llccc@{}}
\toprule
\textbf{Subset} & \textbf{Metric} & \textbf{ReAct} & \textbf{DeReAct} & \textbf{$\Delta$ (pp)} \\
\midrule

\multirow{4}{*}{Overall}
& Evidence complete     & 61.3\% & 69.7\% & $+8.4$ \\
& Constraints satisfied & 82.7\% & 89.1\% & $+6.4$ \\
& Admitted guessing     & 21.0\% & 15.1\% & $-5.9$ \\
& Both criteria met     & 60.1\% & 68.7\% & $+8.6$ \\
\midrule

\multirow{2}{*}{Correct}
& Both criteria met     & 78.4\% & 82.6\% & $+4.2$ \\
& Admitted guessing     &  5.6\% &  5.5\% & $-0.1$ \\
\midrule

\multirow{3}{*}{Incorrect}
& Evidence complete     & 29.8\% & 44.6\% & $+14.8$ \\
& Both criteria met     & 29.3\% & 44.0\% & $+14.7$ \\
& Admitted guessing     & 47.0\% & 32.0\% & $-15.0$ \\

\bottomrule
\end{tabular}
\end{table}

\begin{table}[t]
\centering\small
\setlength{\tabcolsep}{4pt}
\caption{Trajectory grounding on SWE-bench for the Opus Brain, measured by
appropriate testing. Resolved and unresolved subsets are defined by the
benchmark test outcome.}
\label{tab:swe_grounding}

\begin{tabular}{@{}lccc@{}}
\toprule
\textbf{Subset} & \textbf{ReAct} & \textbf{DeReAct} & \textbf{$\Delta$ (pp)} \\
\midrule
Overall    & 78.2\% & 80.8\% & $+2.6$ \\
Resolved   & 80.5\% & 82.3\% & $+1.8$ \\
Unresolved & 73.7\% & 78.0\% & $+4.3$ \\
\bottomrule
\end{tabular}
\end{table}
\section{Conclusion}
\label{sec:conclusion}

We introduced DeReAct, which separates task reasoning from action authorization and completion control. Across GAIA and SWE-bench Verified,
DeReAct improves task success for weaker Brains while remaining comparable
to ReAct with the strongest Brain, where it still produces more grounded
trajectories. Our analyses show that external control is effective when
the underlying policy exhibits the targeted failures and the control policy
is itself sufficient for the task and domain.

These findings suggest a practical design principle for reliable agent
systems: action authorization and completion control need not remain
inside the task-solving policy. Externalizing these decisions provides configurable control points that
improve weaker agents and retain value as base models strengthen. Future work
should validate them across new domains and deployment constraints.

\section{Limitations}
\label{sec:limitations}


Our evaluation is limited to GAIA and SWE-bench Verified and does not cover several practically important agent settings, including multi-turn dialogue, human-in-the-loop workflows, and agents whose actions have real-world side effects. Generalization to these settings remains open, as both failure prevalence and policy sufficiency may change with the interaction structure, available feedback, and deployment constraints.

An important open question is how DeReAct generalizes across model families and
capability regimes. In our experiments, the Critic and Context Manager are
Claude models throughout, and only one open-weight Brain is evaluated. Future
work should test whether the observed capability-dependent behavior and
policy-sufficiency patterns persist across additional model families and
non-Claude gating policies. The configuration procedure in
Section~\ref{subsec:sufficiency} should likewise be validated prospectively on
new Brains and benchmarks.

Grounding is operationalized differently across the two benchmarks: evidence
completeness on GAIA and appropriate testing on SWE-bench Verified. Cross-benchmark
comparisons therefore concern rates of ungrounded completion under
domain-specific criteria rather than identical grounding metrics. All
judge-based analyses use a single judge model (Claude Opus~4.7), and CM
decision-quality ratings are based on 100 sampled trajectories per
configuration.

Cost accounting uses list prices without prompt caching or batching. Because
full-history ReAct benefits more from prefix caching than DeReAct's sliding
windows, the reported cost parity on SWE-bench Verified may change under
caching-aware deployment; cache-aligned windowing remains future work. We also
set $T_{\max}{=}1000$ to study intrinsic termination behavior, which increases
mean cost through a tail of failing runs. Production systems would typically
use tighter step limits, for which we additionally report medians and
percentiles.


\bibliography{references}
\appendix

\clearpage
\appendix
\section{Appendix}

\FloatBarrier

\subsection{Excluded SWE-bench Instances}
\label{app:excluded}

The following eleven SWE-bench Verified instances were excluded from evaluation because their Docker images were unavailable in the official evaluation harness. These instances failed immediately in all runs, across all configurations and attempts, and are therefore removed from the denominator (yielding 489 evaluable instances) identically for every architecture.

\begin{itemize}\itemsep1pt
\item \texttt{astropy\_\_astropy-8707}
\item \texttt{astropy\_\_astropy-8872}
\item \texttt{pylint-dev\_\_pylint-7277}
\item \texttt{scikit-learn\_\_scikit-learn-25102}
\item \texttt{scikit-learn\_\_scikit-learn-25232}
\item \texttt{scikit-learn\_\_scikit-learn-25747}
\item \texttt{scikit-learn\_\_scikit-learn-25931}
\item \texttt{scikit-learn\_\_scikit-learn-25973}
\item \texttt{scikit-learn\_\_scikit-learn-26194}
\item \texttt{scikit-learn\_\_scikit-learn-26323}
\item \texttt{sympy\_\_sympy-20590}
\end{itemize}

\subsection{Extra Tables and Figures}

\begin{figure*}[t]
\centering
\begin{minipage}{0.48\textwidth}
\centering
\includegraphics[width=\textwidth]{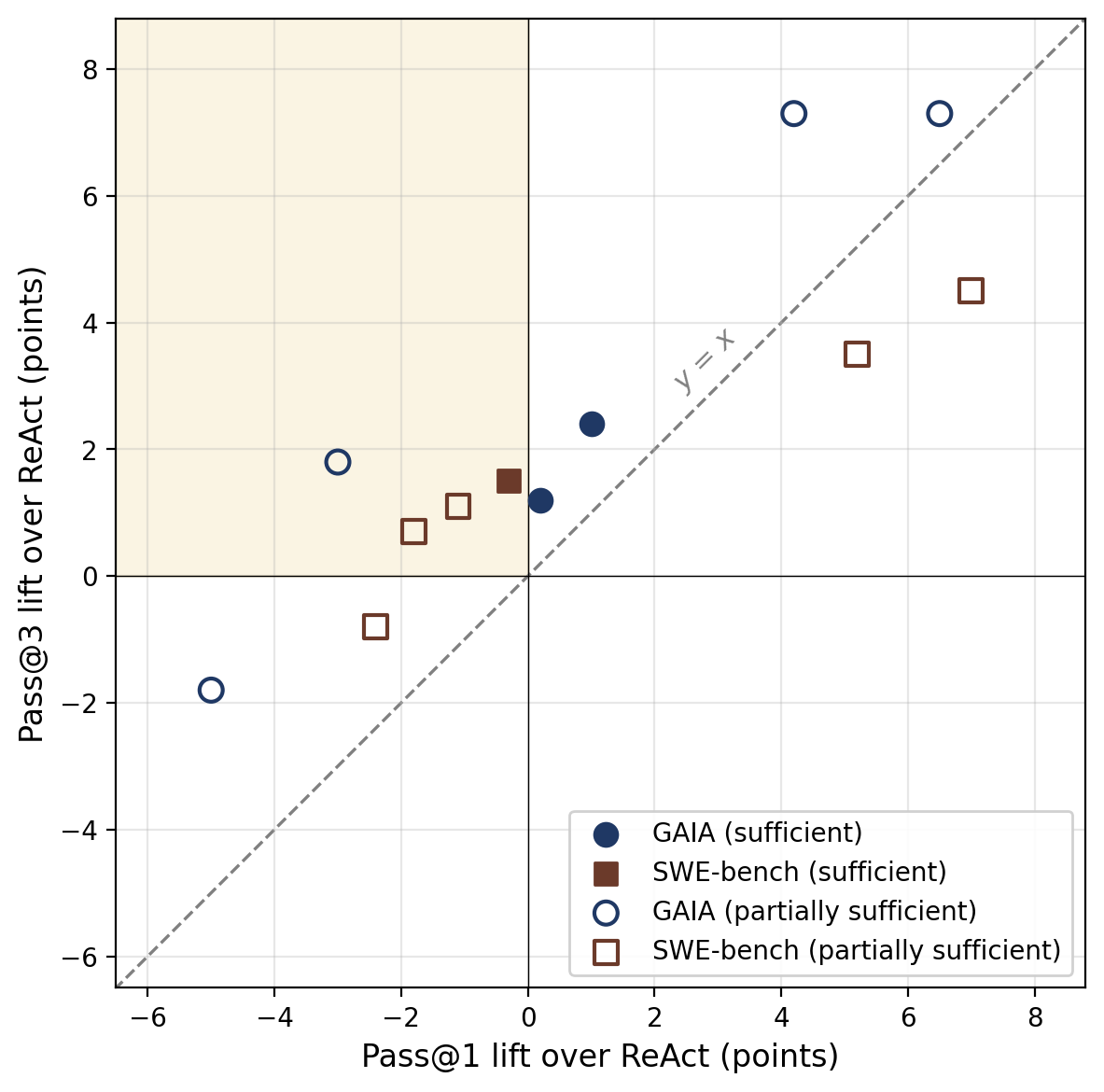}
\caption{Pass@3 lift versus Pass@1 lift over ReAct for all configurations across both benchmarks, Brain tiers, CM models, window sizes, and prompts. The vertical distance from the $y{=}x$ diagonal is Diff. Nearly every point lies above the diagonal. In the shaded region, DeReAct trails ReAct per attempt yet solves more distinct tasks at least once.}
\label{fig:pass3}
\end{minipage}\hfill
\begin{minipage}{0.48\textwidth}
\centering
\includegraphics[width=\textwidth]{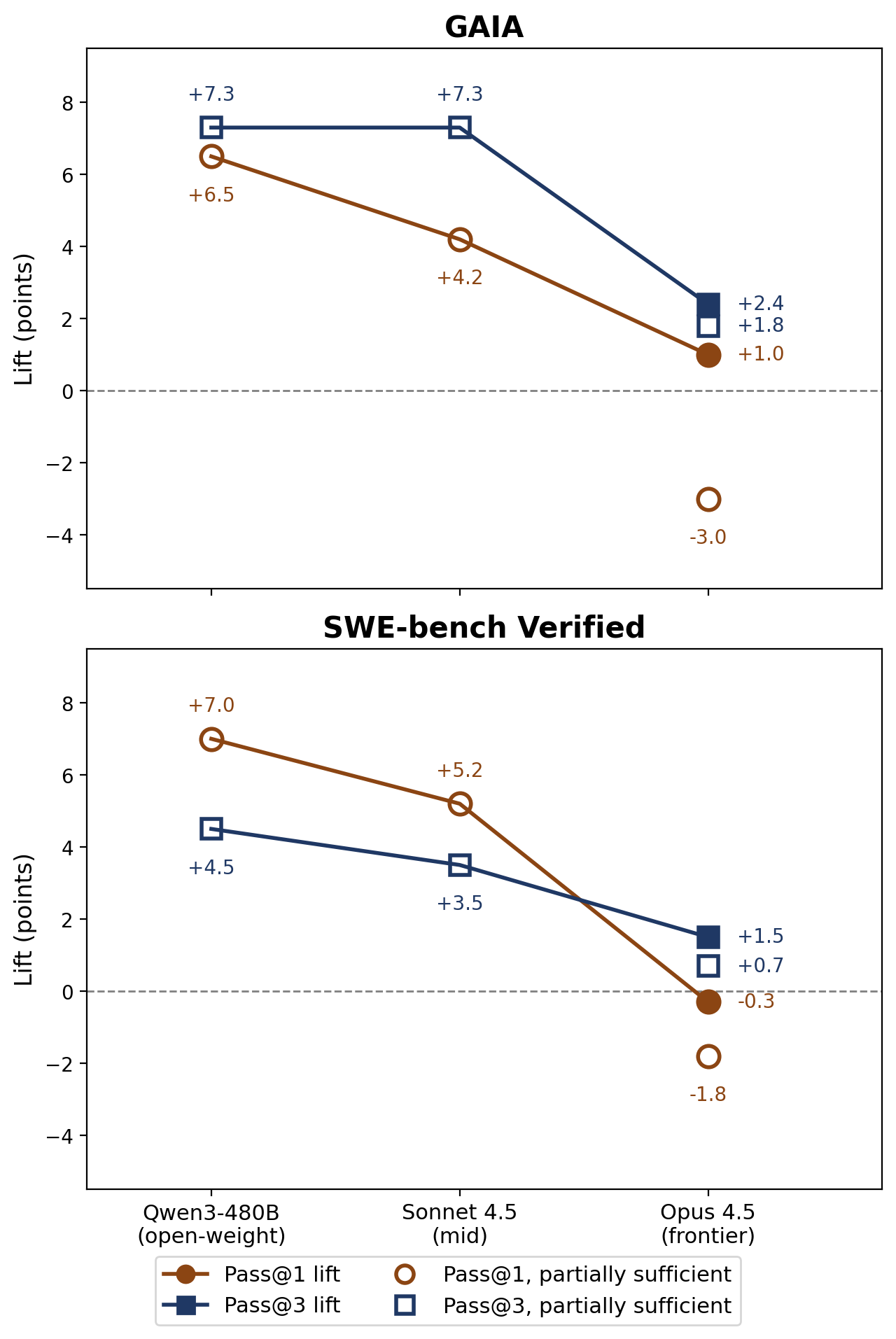}
\caption{Lift over ReAct versus Brain capability, for the sufficient policy configuration at each tier (Qwen, Sonnet: Haiku CM, $W{=}25$; Opus: Opus CM, $W{=}50$). Hollow markers: Opus Brain under the Haiku CM ($W{=}25$). See Table~\ref{tab:compression}.}
\label{fig:compression}
\end{minipage}
\end{figure*}

This subsection collects supplementary tables and figures referenced from the main text. Table~\ref{tab:gaia_react_grounding} reports ReAct's trajectory grounding across all three Brain tiers on GAIA, showing how evidence quality scales with Brain capability. Table~\ref{tab:swe_react_grounding} provides the analogous SWE-bench baseline, measuring appropriate testing rates. These baselines establish the ungrounded-answer rates that the CM's completion veto is designed to reduce. Figure~\ref{fig:pass3} plots Pass@3 lift against Pass@1 lift for all configurations, visualizing the decorrelation effect. Figure~\ref{fig:compression} shows how the DeReAct lift compresses with increasing Brain capability.

\begin{table}[t]
\centering\small
\setlength{\tabcolsep}{2pt}
\caption{GAIA ReAct trajectory grounding across Brain tiers (all three attempts pooled). An independent Opus~4.7 judge, blind to the reference answer, reads each complete trajectory.\protect\footnotemark}
\label{tab:gaia_react_grounding}
\begin{tabular}{@{}lccc@{}}
\toprule
& \textbf{Qwen3-480B} & \textbf{Sonnet 4.5} & \textbf{Opus 4.5} \\ \midrule
Trajectories evaluated        & 490 & 476 & 486 \\
Evidence complete              & 36.7\% & 55.7\% & 61.3\% \\
Constraints satisfied          & 57.3\% & 77.7\% & 82.7\% \\
Admitted guessing              & 37.6\% & 23.1\% & 21.0\% \\
Both criteria met              & 35.5\% & 53.6\% & 60.1\% \\ \midrule
\multicolumn{4}{@{}l}{\emph{Among correct answers}} \\
\quad Evidence complete        & 72.4\% & 75.5\% & 80.0\% \\
\quad Both criteria met        & 69.4\% & 71.9\% & 78.4\% \\
\quad Admitted guessing        & 10.0\% &  7.6\% &  5.6\% \\ \midrule
\multicolumn{4}{@{}l}{\emph{Among incorrect answers}} \\
\quad Evidence complete        & 17.8\% & 33.9\% & 29.8\% \\
\quad Both criteria met        & 17.5\% & 33.5\% & 29.3\% \\
\quad Admitted guessing        & 52.2\% & 40.1\% & 47.0\% \\
\bottomrule
\end{tabular}
\end{table}

\footnotetext{Trajectory counts fall below the theoretical maximum ($3 \times 165 = 495$ for GAIA; $3 \times 489 = 1{,}467$ for SWE-bench) because two exclusion criteria apply.
First, runs that exhausted the step budget ($T_{\max}$) or crashed without producing a final answer (GAIA) or a non-empty code patch (SWE-bench) are excluded, since no agent output exists to judge.
Second, the small number of trajectories for which the Opus~4.7 judge returned an API error or unparseable JSON are excluded from aggregation.
Both criteria filter on structural completeness rather than content quality and are thus plausibly independent of grounding quality; the larger Sonnet shortfall on SWE-bench reflects its higher rate of empty patches (runs completing without producing a diff), consistent with its longer average trajectories hitting the sandbox output timeout.}

\begin{table}[t]
\centering\small
\setlength{\tabcolsep}{4pt}
\caption{SWE-bench ReAct trajectory grounding across Brain tiers (all three attempts pooled). An independent Opus~4.7 judge, blind to the test outcome, assesses whether the agent performed appropriate testing.\protect\footnotemark[\value{footnote}]}
\label{tab:swe_react_grounding}
\begin{tabular}{@{}lccc@{}}
\toprule
& \textbf{Qwen3-480B} & \textbf{Sonnet 4.5} & \textbf{Opus 4.5} \\ \midrule
Trajectories evaluated          & 1417 & 1298 & 1462 \\
Appropriate testing (overall)   &  4.6\% & 80.1\% & 78.2\% \\ \midrule
\multicolumn{4}{@{}l}{\emph{Among resolved (tests passed)}} \\
\quad Appropriate testing       &  8.1\% & 84.5\% & 80.5\% \\ \midrule
\multicolumn{4}{@{}l}{\emph{Among unresolved (tests failed)}} \\
\quad Appropriate testing       &  2.4\% & 74.7\% & 73.7\% \\
\bottomrule
\end{tabular}
\end{table}

\begin{table*}[t]
\centering\small
\setlength{\tabcolsep}{4pt}
\caption{GAIA Policy Activations. Critic Rej. shows percentage of Brain-proposed actions rejected by the Critic. CM Rej. shows percentage of Brain-proposed final-answers rejected by the CM.}
\label{tab:gaia_rej_full}
\begin{tabular}{@{}lllccccc@{}}
\toprule
 &  &  & \multicolumn{2}{c}{\textbf{Critic Rej. (\%)}} & \multicolumn{2}{c}{\textbf{CM Rej. (\%)}} \\ \cmidrule(lr){4-5} \cmidrule(l){6-7}
\textbf{Brain} & \textbf{CM} & \textbf{$W$} & \textbf{DeReAct} & \textbf{Abl.B} & \textbf{DeReAct} & \textbf{Abl.A} \\ \midrule
Qwen3-480B & Haiku & 25 & 28.07 & 4.30 & 93.3 & 92.4 \\
Sonnet 4.5 & Haiku & 25 & 2.20 & 0.16 & 90.4 & 89.2 \\
Opus 4.5 & Haiku & 25 & 2.65 & 0.10 & 81.7 & 85.6 \\
Opus 4.5 & Opus & 25 & 1.25 & 0.10 & 38.5 & 46.1 \\
\midrule
Opus 4.5 & Haiku & 50 & 4.14 & 0.10 & 95.9 & 95.7 \\
Opus 4.5 & Opus & 50 & 1.31 & 0.10 & 55.7 & 54.8 \\
\bottomrule
\end{tabular}
\end{table*}

\begin{table*}[t]
\centering\small
\setlength{\tabcolsep}{4pt}
\caption{GAIA Pass@1 (\%) by difficulty level, mean $\pm$ standard error over three attempts ($W{=}25$, Haiku CM and Critic).}
\label{tab:gaia_levels}
\begin{tabular}{@{}llccc@{}}
\toprule
\textbf{Brain} & \textbf{Config} & \textbf{Level 1} & \textbf{Level 2} & \textbf{Level 3} \\ \midrule
Qwen3-480B & ReAct      & $47.2 \pm 6.2$ & $33.7 \pm 4.3$ & $14.1 \pm 5.6$ \\
           & DeReAct    & $48.4 \pm 6.2$ & $43.0 \pm 4.6$ & $21.8 \pm 6.9$ \\
           & Ablation A & $47.8 \pm 6.3$ & $41.9 \pm 4.4$ & $16.7 \pm 6.2$ \\
           & Ablation B & $46.5 \pm 6.4$ & $34.1 \pm 4.3$ & $17.9 \pm 5.6$ \\ \midrule
Sonnet 4.5 & ReAct      & $55.3 \pm 6.0$ & $55.8 \pm 4.9$ & $21.8 \pm 7.1$ \\
           & DeReAct    & $62.3 \pm 5.9$ & $55.8 \pm 4.6$ & $34.6 \pm 8.6$ \\
           & Ablation A & $60.4 \pm 6.0$ & $58.1 \pm 4.7$ & $32.1 \pm 8.4$ \\
           & Ablation B & $56.0 \pm 6.0$ & $53.5 \pm 4.7$ & $35.9 \pm 8.9$ \\ \midrule
Opus 4.5   & ReAct      & $68.6 \pm 6.0$ & $65.5 \pm 4.6$ & $34.6 \pm 8.4$ \\
           & DeReAct    & $66.0 \pm 5.6$ & $61.6 \pm 4.4$ & $33.3 \pm 8.1$ \\
           & Ablation A & $66.0 \pm 5.6$ & $60.5 \pm 4.6$ & $34.6 \pm 8.8$ \\
           & Ablation B & $64.8 \pm 6.2$ & $64.3 \pm 5.0$ & $32.1 \pm 8.8$ \\
\bottomrule
\end{tabular}
\end{table*}

\begin{table*}[t]
\centering\small
\setlength{\tabcolsep}{4pt}
\caption{GAIA Pass@1 accuracy (\%), averaged over three attempts. Lifts are relative to the standard ReAct baseline for the corresponding W at the same tier. Values are mean $\pm$ standard error. $p$-value from paired permutation test. (GAIA levels breakdown: Table~\ref{tab:gaia_levels})}
\label{tab:gaia}
\begin{tabular}{@{}llccccccc@{}}
\toprule
\textbf{Brain} & \textbf{CM} & \textbf{W} & \textbf{DeReAct} & \textbf{ReAct} & \textbf{Abl.A} & \textbf{Abl.B} & \textbf{Lift} & \textbf{$p$} \\ \midrule
Qwen3-480B & Haiku & 25 & $41.4 \pm 3.3$ & $34.9 \pm 3.2$ & $39.8 \pm 3.1$ & $35.6 \pm 3.4$ & $+6.5$ & .010 \\
Sonnet 4.5 & Haiku & 25 & $54.5 \pm 3.4$ & $50.3 \pm 3.5$ & $54.7 \pm 3.3$ & $51.5 \pm 3.2$ & $+4.2$ & .170 \\
Opus 4.5 & Haiku & 25 & $58.6 \pm 3.3$ & $61.6 \pm 3.5$ & $58.2 \pm 3.2$ & $59.4 \pm 3.3$ & $-3.0$ & .201 \\ \midrule
Opus 4.5 & Opus & 25 & $61.8 \pm 3.4$ & $61.6 \pm 3.5$ & $61.2 \pm 3.3$ & $59.4 \pm 3.3$ & $+0.2$ & 1.000 \\
Opus 4.5 & Haiku & 50 & $56.6 \pm 3.4$ & $61.6 \pm 3.5$ & $59.6 \pm 3.4$ & $59.4 \pm 3.3$ & $-5.0$ & .021 \\
Opus 4.5 & Opus & 50 & $62.6 \pm 3.4$ & $61.6 \pm 3.5$ & $62.6 \pm 3.3$ & $59.4 \pm 3.3$ & $+1.0$ & .697 \\
\bottomrule
\end{tabular}
\end{table*}

\begin{table*}[t]
\centering\small
\setlength{\tabcolsep}{4pt}
\caption{GAIA Pass@3 accuracy (\%), measuring the percentage of questions answered correctly in at least 1 of 3 attempts. Lifts are relative to the standard ReAct baseline for the corresponding W at the same tier. Values are mean $\pm$ standard error. $p$-value from paired permutation test.}
\label{tab:gaia-pass3}
\begin{tabular}{@{}llccccccc@{}}
\toprule
\textbf{Brain} & \textbf{CM} & \textbf{W} & \textbf{DeReAct} & \textbf{ReAct} & \textbf{Abl.A} & \textbf{Abl.B} & \textbf{Lift} & \textbf{$p$} \\ \midrule
Qwen3-480B & Haiku & 25 & $55.2 \pm 3.9$ & $47.9 \pm 3.9$ & $54.5 \pm 3.9$ & $48.5 \pm 3.9$ & $+7.3$ & .068 \\
Sonnet 4.5 & Haiku & 25 & $68.5 \pm 3.6$ & $61.2 \pm 3.8$ & $67.3 \pm 3.7$ & $63.6 \pm 3.7$ & $+7.3$ & .060 \\
Opus 4.5 & Haiku & 25 & $72.7 \pm 3.5$ & $70.9 \pm 3.5$ & $70.3 \pm 3.6$ & $64.8 \pm 3.7$ & $+1.8$ & .636 \\ \midrule
Opus 4.5 & Opus & 25 & $72.1 \pm 3.5$ & $70.9 \pm 3.5$ & $71.5 \pm 3.5$ & $64.8 \pm 3.7$ & $+1.2$ & .809 \\
Opus 4.5 & Haiku & 50 & $69.1 \pm 3.6$ & $70.9 \pm 3.5$ & $72.1 \pm 3.5$ & $64.8 \pm 3.7$ & $-1.8$ & .626 \\
Opus 4.5 & Opus & 50 & $73.3 \pm 3.4$ & $70.9 \pm 3.5$ & $71.5 \pm 3.5$ & $64.8 \pm 3.7$ & $+2.4$ & .506 \\
\bottomrule
\end{tabular}
\end{table*}

\begin{table*}[t]
\centering\small
\setlength{\tabcolsep}{3.5pt}
\caption{GAIA cost and step distributions per run (USD, list prices, no caching). Means are dominated by a tail of near-limit trajectories; medians characterize typical runs.}
\label{tab:gaia_cost}
\begin{tabular}{@{}lllccccccc@{}}
\toprule
& & & \multicolumn{4}{c}{\textbf{Cost (\$)}} & \multicolumn{2}{c}{\textbf{Steps}} \\
\cmidrule(lr){4-7}\cmidrule(l){8-9}
\textbf{Brain / CM / $W$} & \textbf{Config} & & \textbf{Mean} & \textbf{Median} & \textbf{P90} & \textbf{P95} & \textbf{Median} & \textbf{P95} \\ \midrule
Qwen3-480B / Haiku / 25 & ReAct   & & 0.02 & 0.01 & 0.04 & 0.07 & 7 & 20 \\
                        & DeReAct & & 3.67 & 0.35 & 14.06 & 17.31 & 28 & 1000 \\
                        & Abl.\ A & & 4.32 & 0.27 & 16.43 & 20.45 & 26 & 1000 \\
                        & Abl.\ B & & 0.07 & 0.03 & 0.16 & 0.27 & 8 & 20 \\ \midrule
Sonnet 4.5 / Haiku / 25 & ReAct   & & 0.57 & 0.13 & 1.64 & 2.48 & 10 & 51 \\
                        & DeReAct & & 5.08 & 0.43 & 12.25 & 32.74 & 16 & 510 \\
                        & Abl.\ A & & 4.66 & 0.38 & 10.87 & 27.76 & 15 & 460 \\
                        & Abl.\ B & & 0.68 & 0.15 & 1.95 & 2.89 & 10 & 48 \\ \midrule
Opus 4.5 / Haiku / 25   & ReAct   & & 1.70 & 0.21 & 5.17 & 8.35 & 10 & 74 \\
                        & DeReAct & & 7.17 & 0.47 & 11.93 & 43.63 & 13 & 359 \\
                        & Abl.\ A & & 7.71 & 0.52 & 17.37 & 48.49 & 14 & 468 \\
                        & Abl.\ B & & 2.11 & 0.24 & 6.94 & 9.77 & 10 & 76 \\ \midrule
Opus 4.5 / Opus / 25    & DeReAct & & 5.05 & 0.59 & 14.36 & 22.62 & 10 & 192 \\
                        & Abl.\ A & & 5.20 & 0.54 & 14.40 & 30.99 & 10 & 255 \\
                        & Abl.\ B & & 3.50 & 0.40 & 11.45 & 16.27 & 10 & 76 \\ \midrule
Opus 4.5 / Haiku / 50   & DeReAct & & 9.66 & 0.41 & 23.41 & 65.32 & 12 & 452 \\
                        & Abl.\ A & & 8.84 & 0.55 & 22.20 & 55.83 & 15 & 410 \\ \midrule
Opus 4.5 / Opus / 50    & DeReAct & & 4.85 & 0.64 & 14.61 & 23.48 & 11 & 141 \\
                        & Abl.\ A & & 4.14 & 0.53 & 11.33 & 21.42 & 11 & 125 \\
\bottomrule
\end{tabular}
\end{table*}

\begin{table*}[t]
\centering\small
\setlength{\tabcolsep}{4pt}
\caption{GAIA veto yield. \textbf{Traj.\ Rej.}: fraction of trajectories with $\geq 1$ CM final-answer rejection. \textbf{Res.\ $\mid$ Rej.}: resolution rate among rejection trajectories. \textbf{ReAct all}: ReAct's resolution rate on the same (instance, attempt) pairs. \textbf{Yield} $=$ Res.\ $\mid$ Rej.\ $-$ ReAct all. \textbf{ReAct (fail)} / \textbf{Evid.\ (fail)}: on pairs the configuration failed, ReAct's resolution rate and the evidence-complete fraction of those wins (raw counts small; indicative).}
\label{tab:gaia_veto_yield}
\begin{tabular}{@{}lll cccc cc@{}}
\toprule
& & & \multicolumn{4}{c}{\textbf{Veto yield}} & \multicolumn{2}{c}{\textbf{Pass@1 parity}} \\
\cmidrule(lr){4-7}\cmidrule(l){8-9}
\textbf{Brain} & \textbf{CM} & \textbf{$W$} & \textbf{Traj.\ Rej.} & \textbf{Res.\ $\mid$ Rej.} & \textbf{ReAct all} & \textbf{Yield} & \textbf{ReAct (fail)} & \textbf{Evid.\ (fail)} \\ \midrule
\multicolumn{9}{@{}l}{\emph{DeReAct}} \\
Qwen3-480B & Haiku & 25 & 65.5 & 30.2 & 23.1 & $+7.1$ & 11.5 & 55 \\
Sonnet 4.5 & Haiku & 25 & 40.8 & 43.1 & 35.6 & $+7.5$ & 23.5 & 44 \\
Opus 4.5   & Haiku & 25 & 41.8 & 51.7 & 52.7 & $-1.0$ & 23.0 & 33 \\
Opus 4.5   & Opus  & 25 & 22.6 & 39.3 & 38.4 & $+0.9$ & 14.7 & 67 \\
Opus 4.5   & Haiku & 50 & 40.4 & 44.0 & 48.0 & $-4.0$ & 20.5 & 58 \\
Opus 4.5   & Opus  & 50 & 29.9 & 39.2 & 35.8 & $+3.4$ & 15.6 & 20 \\ \midrule
\multicolumn{9}{@{}l}{\emph{Ablation A (CM, no Critic)}} \\
Qwen3-480B & Haiku & 25 & 66.9 & 29.0 & 22.7 & $+6.3$ & 8.5  & 27 \\
Sonnet 4.5 & Haiku & 25 & 39.2 & 41.8 & 31.4 & $+10.4$ & 16.8 & 50 \\
Opus 4.5   & Haiku & 25 & 47.3 & 47.0 & 50.0 & $-3.0$ & 22.6 & 45 \\
Opus 4.5   & Opus  & 25 & 29.9 & 39.9 & 35.8 & $+4.1$ & 13.5 & 67 \\
Opus 4.5   & Haiku & 50 & 50.1 & 48.8 & 49.6 & $-0.8$ & 22.8 & 50 \\
Opus 4.5   & Opus  & 50 & 31.1 & 37.0 & 35.1 & $+1.9$ & 14.4 & 67 \\
\bottomrule
\end{tabular}
\end{table*}

\begin{table*}[t]
\centering\small
\setlength{\tabcolsep}{4pt}
\caption{SWE-bench Verified Pass@1 resolution (\%, averaged over three attempts, \emph{base} CM prompt (\ref{app:prompts:cm:swe})). Lifts are relative to ReAct. Values are mean $\pm$ standard error. $p$-value from paired permutation test.}
\label{tab:swe}
\begin{tabular}{@{}llccccccc@{}}
\toprule
\textbf{Brain} & \textbf{CM} & \textbf{W} & \textbf{DeReAct} & \textbf{ReAct} & \textbf{Abl.A} & \textbf{Abl.B} & \textbf{Lift} & \textbf{$p$} \\ \midrule
Qwen3-480B & Haiku & 25 & $42.0 \pm 0.7$ & $37.0 \pm 0.7$ & $37.6 \pm 0.2$ & $43.6 \pm 0.2$ & $+5.0$ & ${<}.001$ \\
Sonnet~4.5 & Haiku & 25 & $52.7 \pm 0.4$ & $48.9 \pm 0.8$ & $53.0 \pm 0.0$ & $51.3 \pm 0.9$ & $+3.8$ & .008 \\
Opus~4.5 & Haiku & 25 & $61.2 \pm 1.0$ & $65.9 \pm 0.5$ & $63.9 \pm 1.0$ & $66.1 \pm 0.2$ & $-4.7$ & ${<}.001$ \\
\bottomrule
\end{tabular}
\end{table*}

\begin{table*}[t]
\centering\small
\setlength{\tabcolsep}{4pt}
\caption{SWE-bench Verified Pass@1 resolution (\%, averaged over three attempts, \emph{improved} CM prompt (\ref{app:prompts:cm:swe})  ). Lifts are relative to ReAct. Values are mean $\pm$ standard error. $p$-value from paired permutation test.}
\label{tab:swe_improved}
\begin{tabular}{@{}llccccccc@{}}
\toprule
\textbf{Brain} & \textbf{CM} & \textbf{W} & \textbf{DeReAct} & \textbf{ReAct} & \textbf{Abl.A} & \textbf{Abl.B} & \textbf{Lift} & \textbf{$p$} \\ \midrule
Qwen3-480B & Haiku & 25 & $44.0 \pm 1.8$ & $37.0 \pm 0.7$ & $33.5 \pm 0.5$ & $43.6 \pm 0.2$ & $+7.0$ & ${<}.001$ \\
Sonnet~4.5 & Haiku & 25 & $54.1 \pm 0.4$ & $48.9 \pm 0.8$ & $52.2 \pm 0.5$ & $51.3 \pm 0.9$ & $+5.2$ & ${<}.001$ \\
Opus~4.5 & Haiku & 25 & $64.1 \pm 0.2$ & $65.9 \pm 0.5$ & $64.3 \pm 0.9$ & $66.1 \pm 0.2$ & $-1.8$ & .092 \\ 
Opus~4.5 & Opus & 25 & $63.5 \pm 1.0$ & $65.9 \pm 0.5$ & $64.7 \pm 1.0$ & $66.1 \pm 0.2$ & $-2.4$ & .020 \\ \midrule
Opus~4.5 & Haiku & 50 & $64.8 \pm 0.5$ & $65.9 \pm 0.5$ & $64.6 \pm 0.3$ & $66.1 \pm 0.2$ & $-1.1$ & .250 \\
Opus~4.5 & Opus & 50 & $65.6 \pm 0.8$ & $65.9 \pm 0.5$ & $65.5 \pm 0.3$ & $66.1 \pm 0.2$ & $-0.3$ & .753 \\
\bottomrule
\end{tabular}
\end{table*}

\begin{table*}[t]
\centering\small
\setlength{\tabcolsep}{4pt}
\caption{SWE-bench Verified Pass@3 resolution (\%, \emph{base} CM prompt (\ref{app:prompts:cm:swe})). Pass@3 measures the percentage of instances resolved in at least 1 of 3 attempts. Lifts are relative to ReAct. Values are mean $\pm$ standard error. $p$-value from paired permutation test.}
\label{tab:swe-pass3}
\begin{tabular}{@{}llccccccc@{}}
\toprule
\textbf{Brain} & \textbf{CM} & \textbf{W} & \textbf{DeReAct} & \textbf{ReAct} & \textbf{Abl.A} & \textbf{Abl.B} & \textbf{Lift} & \textbf{$p$} \\ \midrule
Qwen3-480B & Haiku & 25 & $58.7 \pm 2.2$ & $54.2 \pm 2.3$ & $53.8 \pm 2.3$ & $57.7 \pm 2.2$ & $+4.5$ & .016 \\
Sonnet~4.5 & Haiku & 25 & $63.8 \pm 2.2$ & $62.8 \pm 2.2$ & $65.8 \pm 2.1$ & $64.4 \pm 2.2$ & $+1.0$ & .586 \\
Opus~4.5 & Haiku & 25 & $70.1 \pm 2.1$ & $70.1 \pm 2.1$ & $70.8 \pm 2.1$ & $70.8 \pm 2.1$ & $+0.0$ & 1.000 \\
\bottomrule
\end{tabular}
\end{table*}

\begin{table*}[t]
\centering\small
\setlength{\tabcolsep}{4pt}
\caption{SWE-bench Verified Pass@3 resolution (\%, \emph{improved} CM prompt (\ref{app:prompts:cm:swe})). Pass@3 measures the percentage of instances resolved in at least 1 of 3 attempts. Lifts are relative to ReAct. Values are mean $\pm$ standard error. $p$-value from paired permutation test.}
\label{tab:swe_improved-pass3}
\begin{tabular}{@{}llccccccc@{}}
\toprule
\textbf{Brain} & \textbf{CM} & \textbf{W} & \textbf{DeReAct} & \textbf{ReAct} & \textbf{Abl.A} & \textbf{Abl.B} & \textbf{Lift} & \textbf{$p$} \\ \midrule
Qwen3-480B & Haiku & 25 & $58.7 \pm 2.2$ & $54.2 \pm 2.3$ & $53.8 \pm 2.3$ & $57.7 \pm 2.2$ & $+4.5$ & .014 \\
Sonnet~4.5 & Haiku & 25 & $66.3 \pm 2.1$ & $62.8 \pm 2.2$ & $65.2 \pm 2.2$ & $64.4 \pm 2.2$ & $+3.5$ & .039 \\
Opus~4.5 & Haiku & 25 & $70.8 \pm 2.1$ & $70.1 \pm 2.1$ & $70.1 \pm 2.1$ & $70.8 \pm 2.1$ & $+0.7$ & .725 \\
Opus~4.5 & Opus & 25 & $69.3 \pm 2.1$ & $70.1 \pm 2.1$ & $70.8 \pm 2.1$ & $70.8 \pm 2.1$ & $-0.8$ & .600 \\ \midrule
Opus~4.5 & Haiku & 50 & $71.2 \pm 2.0$ & $70.1 \pm 2.1$ & $70.1 \pm 2.1$ & $70.8 \pm 2.1$ & $+1.1$ & .480 \\
Opus~4.5 & Opus & 50 & $71.6 \pm 2.0$ & $70.1 \pm 2.1$ & $71.8 \pm 2.0$ & $70.8 \pm 2.1$ & $+1.5$ & .269 \\
\bottomrule
\end{tabular}
\end{table*}

\begin{table*}[t]
\centering\small
\setlength{\tabcolsep}{4pt}
\caption{SWE-bench Policy Activations. Critic Rej. shows percentage of Brain-proposed actions rejected by the Critic. CM Rej. shows percentage of Brain-proposed final-answers rejected by the CM.}
\label{tab:swe_rej_full}
\begin{tabular}{@{}lllccccc@{}}
\toprule
 &  &  & \multicolumn{2}{c}{\textbf{Critic Rej. (\%)}} & \multicolumn{2}{c}{\textbf{CM Rej. (\%)}} \\ \cmidrule(lr){4-5} \cmidrule(l){6-7}
\textbf{Brain} & \textbf{CM} & \textbf{$W$} & \textbf{DeReAct} & \textbf{Abl.B} & \textbf{DeReAct} & \textbf{Abl.A} \\ \midrule
Qwen3-480B & Haiku (Base Prompt) & 25 & 22.87 & 32.55 & 71.42 & 77.39 \\
Sonnet 4.5 & Haiku (Base Prompt) & 25 & 2.08 & 1.67 & 42.91 & 21.76 \\
Opus 4.5 & Haiku (Base Prompt) & 25 & 2.39 & 2.01 & 10.18 & 8.95 \\
Qwen3-480B & Haiku & 25 & 16.15 & 32.55 & 93.09 & 93.08 \\
Sonnet 4.5 & Haiku & 25 & 2.06 & 1.67 & 72.36 & 65.29 \\
Opus 4.5 & Haiku & 25 & 2.08 & 2.01 & 56.48 & 56.38 \\
Opus 4.5 & Opus & 25 & 2.15 & 2.01 & 19.63 & 19.85 \\
\midrule \\
Opus 4.5 & Haiku & 50 & 2.19 & 2.01 & 59.27 & 55.37 \\
Opus 4.5 & Opus & 50 & 2.18 & 2.01 & 20.09 & 20.66 \\
\bottomrule
\end{tabular}
\end{table*}

\begin{table*}[t]
\centering\small
\setlength{\tabcolsep}{3.5pt}
\caption{SWE-bench cost and step distributions per run (USD, list prices, no caching). Mean/median gaps are far smaller than on GAIA because the test harness supplies a natural termination signal; under the Opus Brain with a matched CM, DeReAct reaches mean-cost parity with ReAct.}
\label{tab:swe_cost}
\begin{tabular}{@{}lllccccccc@{}}
\toprule
& & & \multicolumn{4}{c}{\textbf{Cost (\$)}} & \multicolumn{2}{c}{\textbf{Steps}} \\
\cmidrule(lr){4-7}\cmidrule(l){8-9}
\textbf{Brain / CM / $W$} & \textbf{Config} & & \textbf{Mean} & \textbf{Median} & \textbf{P90} & \textbf{P95} & \textbf{Median} & \textbf{P95} \\ \midrule
Qwen3-480B / Haiku / 25 & ReAct & & 0.32 & 0.13 & 0.55 & 0.87 & 34 & 98 \\
                        & DeReAct & & 2.79 & 1.63 & 7.16 & 10.23 & 154 & 1000 \\
                        & Abl.\ B & & 1.95 & 0.50 & 1.97 & 3.05 & 33 & 93 \\ \midrule
Sonnet 4.5 / Haiku / 25 & ReAct   & & 3.55 & 2.91 & 6.84 & 8.72 & 52 & 100 \\
                        & DeReAct & & 5.38 & 2.82 & 11.66 & 19.41 & 68 & 394 \\
                        & Abl.\ A & & 4.92 & 2.63 & 10.37 & 16.41 & 65 & 344 \\
                        & Abl.\ B & & 4.58 & 3.72 & 9.17 & 11.96 & 51 & 100 \\ \midrule
Opus 4.5 / Haiku / 25   & ReAct   & & 3.94 & 2.60 & 8.08 & 10.89 & 38 & 82 \\
                        & DeReAct & & 4.12 & 2.49 & 8.66 & 13.65 & 42 & 167 \\
                        & Abl.\ A & & 4.13 & 2.56 & 9.12 & 13.52 & 44 & 173 \\
                        & Abl.\ B & & 4.59 & 2.97 & 9.06 & 13.09 & 37 & 80 \\ \midrule
Opus 4.5 / Opus / 25    & DeReAct & & 3.58 & 2.46 & 7.04 & 9.21 & 36 & 100 \\
                        & Abl.\ A & & 3.67 & 2.46 & 7.02 & 9.75 & 36 & 105 \\ \midrule
Opus 4.5 / Haiku / 50   & DeReAct & & 4.84 & 2.78 & 10.59 & 15.48 & 41 & 118 \\
                        & Abl.\ A & & 4.65 & 2.78 & 10.45 & 14.12 & 43 & 116 \\ \midrule
Opus 4.5 / Opus / 50    & DeReAct & & 3.94 & 2.49 & 8.13 & 11.38 & 35 & 88 \\
                        & Abl.\ A & & 3.83 & 2.63 & 7.84 & 10.67 & 37 & 86 \\
\bottomrule
\end{tabular}
\end{table*}

\begin{table*}[t]
\centering\small
\setlength{\tabcolsep}{4pt}
\caption{SWE-bench veto yield. Columns as in Table~\ref{tab:gaia_veto_yield}, with evidence assessed as appropriate testing.}
\label{tab:swe_veto_yield}
\begin{tabular}{@{}llll cccc cc@{}}
\toprule
& & & & \multicolumn{4}{c}{\textbf{Veto yield}} & \multicolumn{2}{c}{\textbf{Pass@1 parity}} \\
\cmidrule(lr){5-8}\cmidrule(l){9-10}
\textbf{Brain} & \textbf{CM} & \textbf{Prompt} & \textbf{$W$} & \textbf{Traj.\ Rej.} & \textbf{Res.\ $\mid$ Rej.} & \textbf{ReAct all} & \textbf{Yield} & \textbf{ReAct (fail)} & \textbf{Evid.\ (fail)} \\ \midrule
\multicolumn{10}{@{}l}{\emph{DeReAct}} \\
Qwen3-480B & Haiku & base  & 25 & 72.3 & 41.1 & 35.3 & $+5.8$ & 14.9 & 5.4  \\
Qwen3-480B & Haiku & impr. & 25 & 96.3 & 43.7 & 37.3 & $+6.4$ & 17.9 & 9.2  \\
Sonnet 4.5 & Haiku & base  & 25 & 3.4  & 32.0 & 26.0 & $+6.0$ & 14.7 & 20.0 \\
Sonnet 4.5 & Haiku & impr. & 25 & 30.3 & 48.9 & 39.2 & $+9.7$ & 17.6 & 70.0 \\
Opus 4.5   & Haiku & base  & 25 & 9.3  & 53.3 & 60.6 & $-7.3$ & 28.1 & 72.2 \\
Opus 4.5   & Haiku & impr. & 25 & 49.4 & 59.2 & 59.9 & $-0.7$ & 14.9 & 63.6 \\
Opus 4.5   & Opus  & impr. & 25 & 18.3 & 63.4 & 67.2 & $-3.8$ & 25.5 & 44.0 \\
Opus 4.5   & Haiku & impr. & 50 & 51.7 & 57.4 & 58.4 & $-1.0$ & 13.9 & 66.7 \\
Opus 4.5   & Opus  & impr. & 50 & 19.8 & 59.3 & 61.0 & $-1.7$ & 17.8 & 61.9 \\ \midrule
\multicolumn{10}{@{}l}{\emph{Ablation A (CM, no Critic)}} \\
Qwen3-480B & Haiku & base  & 25 & 75.3 & 34.2 & 33.9 & $+0.3$ & 16.9 & 4.9  \\
Qwen3-480B & Haiku & impr. & 25 & 95.6 & 33.4 & 37.0 & $-3.6$ & 23.6 & 5.4  \\
Sonnet 4.5 & Haiku & base  & 25 & 2.4  & 40.0 & 34.3 & $+5.7$ & 14.3 & 66.7 \\
Sonnet 4.5 & Haiku & impr. & 25 & 31.0 & 48.4 & 41.5 & $+6.9$ & 16.2 & 65.8 \\
Opus 4.5   & Haiku & base  & 25 & 8.0  & 55.1 & 61.9 & $-6.8$ & 24.5 & 53.8 \\
Opus 4.5   & Haiku & impr. & 25 & 50.9 & 59.5 & 60.5 & $-1.0$ & 15.2 & 65.2 \\
Opus 4.5   & Opus  & impr. & 25 & 18.1 & 64.5 & 65.7 & $-1.2$ & 18.1 & 58.8 \\
Opus 4.5   & Haiku & impr. & 50 & 50.9 & 58.8 & 59.9 & $-1.1$ & 14.7 & 64.4 \\
Opus 4.5   & Opus  & impr. & 50 & 20.3 & 58.9 & 59.3 & $-0.4$ & 18.0 & 59.1 \\
\bottomrule
\end{tabular}
\end{table*}

\begin{table*}[t]
\centering\small
\setlength{\tabcolsep}{4pt}
\caption{Context Manager decision quality, rated by an independent Opus~4.7 judge reading each complete trajectory (100 sampled per config). \textsc{State} synthesis improves with CM model scale on both benchmarks; \textsc{Final Answer validation (FAV)} does not track scale and instead tracks the domain and prompt.}
\label{tab:cm_quality}
\begin{tabular}{@{}lllcccc@{}}
\toprule
& & & \multicolumn{2}{c}{\textbf{\textsc{State}}} & \multicolumn{2}{c}{\textbf{FAV}} \\
\cmidrule(lr){4-5}\cmidrule(lr){6-7}
\textbf{Bench} & \textbf{Config} & \textbf{Brain / CM / Prompt / $W$} & \textbf{GOOD} & \textbf{POOR} & \textbf{GOOD} & \textbf{POOR} \\ \midrule
\multirow{12}{*}{GAIA} & \multirow{6}{*}{DeReAct}
  & Qwen / Haiku / impr.\ / 25  & 69.7\% & 0.0\% & 36.0\% & 31.5\% \\
& & Sonnet / Haiku / impr.\ / 25 & 80.7\% & 1.1\% & 53.4\% & 17.0\% \\
& & Opus / Haiku / impr.\ / 25   & 80.2\% & 1.1\% & 48.4\% & 19.8\% \\
& & Opus / Opus / impr.\ / 25    & 93.9\% & 1.0\% & 41.4\% & 22.2\% \\
& & Opus / Haiku / impr.\ / 50   & 79.6\% & 1.1\% & 44.1\% & 25.8\% \\
& & Opus / Opus / impr.\ / 50    & 97.0\% & 0.0\% & 54.0\% & 16.0\% \\ \cmidrule(l){2-7}
& \multirow{6}{*}{Abl.\ A}
  & Qwen / Haiku / impr.\ / 25  & 52.7\% & 2.2\% & 29.0\% & 46.2\% \\
& & Sonnet / Haiku / impr.\ / 25 & 86.0\% & 0.0\% & 44.2\% & 10.5\% \\
& & Opus / Haiku / impr.\ / 25   & 87.9\% & 1.1\% & 42.9\% & 12.1\% \\
& & Opus / Opus / impr.\ / 25    & 94.9\% & 2.0\% & 40.4\% & 13.1\% \\
& & Opus / Haiku / impr.\ / 50   & 84.4\% & 0.0\% & 49.0\% & 15.6\% \\
& & Opus / Opus / impr.\ / 50    & 98.0\% & 1.0\% & 48.0\% & 6.0\% \\ \midrule
\multirow{17}{*}{SWE} & \multirow{9}{*}{DeReAct}
  & Qwen / Haiku / base / 25    & 51.0\% & 4.0\% & 20.0\% & 34.0\% \\
& & Qwen / Haiku / impr.\ / 25   & 37.0\% & 3.0\% & 37.0\% & 33.0\% \\
& & Sonnet / Haiku / base / 25   & 64.0\% & 7.0\% & 39.0\% & 16.0\% \\
& & Sonnet / Haiku / impr.\ / 25 & 70.0\% & 5.0\% & 50.0\% & 17.0\% \\
& & Opus / Haiku / base / 25     & 83.0\% & 0.0\% & 55.0\% & 8.0\% \\
& & Opus / Haiku / impr.\ / 25   & 88.0\% & 0.0\% & 72.0\% & 2.0\% \\
& & Opus / Opus / impr.\ / 25    & 94.0\% & 0.0\% & 69.0\% & 0.0\% \\
& & Opus / Haiku / impr.\ / 50   & 83.0\% & 0.0\% & 65.0\% & 3.0\% \\
& & Opus / Opus / impr.\ / 50    & 97.0\% & 0.0\% & 69.0\% & 3.0\% \\ \cmidrule(l){2-7}
& \multirow{9}{*}{Abl.\ A}
  & Qwen / Haiku / base / 25    & 49.0\% & 4.0\% & 19.0\% & 28.0\% \\
& & Qwen / Haiku / impr.\ / 25   & 27.6\% & 6.1\% & 36.7\% & 32.7\% \\
& & Sonnet / Haiku / base / 25   & 70.0\% & 5.0\% & 39.0\% & 23.0\% \\
& & Sonnet / Haiku / impr.\ / 25 & 68.0\% & 4.0\% & 44.0\% & 18.0\% \\
& & Opus / Haiku / base / 25     & 88.0\% & 1.0\% & 54.0\% & 6.0\% \\
& & Opus / Haiku / impr.\ / 25   & 95.0\% & 0.0\% & 75.0\% & 2.0\% \\
& & Opus / Opus / impr.\ / 25    & 94.0\% & 0.0\% & 63.0\% & 3.0\% \\
& & Opus / Haiku / impr.\ / 50   & 86.0\% & 0.0\% & 65.0\% & 6.0\% \\
& & Opus / Opus / impr.\ / 50    & 93.0\% & 0.0\% & 64.0\% & 4.0\% \\
\bottomrule
\end{tabular}
\end{table*}

\clearpage
\onecolumn

\FloatBarrier
\section{Prompt Templates}
\label{app:prompts}

This appendix presents the prompts used in our evaluation. The ReAct system prompt is universal: identical across all benchmarks (GAIA and SWE-bench Verified) and across all configurations (Full DeReAct, Brain + Context Manager, Brain + Critic, ReAct). The Critic and Context Manager prompts adapt per benchmark.

\subsection{ReAct Prompt (Universal)}
\label{app:prompts:react}

The same system prompt is used for ReAct (the monolithic baseline) and for the Brain component in all DeReAct configurations across all benchmarks. This normalization is deliberate: holding the Brain's reasoning prompt constant ensures that any architectural effect on Brain behavior is attributable to the architectural injection of the Critic and Context Manager, not to differences in how the Brain itself is prompted.

\begin{promptbox}{System Prompt: ReAct \& Brain Component}
You are an AI agent using the ReAct (Reasoning + Acting) framework to accomplish tasks.\\
 
For each step, you should:\\
1. Think about what needs to be done next (Thought)\\
2. Decide on a single action to take (Action)\\
3. Execute the action and observe the result (Observation)\\
 
You MUST respond with exactly ONE Thought and ONE Action per response.\\
Do NOT plan multiple steps ahead.\\
 
Format your response as:\\
Thought: [your reasoning about what to do next]\\
Action: [the command to execute]\\
 
When the task is complete, respond with:\\
Thought: [explanation of why task is complete]\\
Final Answer: [summary of what was accomplished]\\
 
"Action" and "Final Answer" should be mutually exclusive, and therefore they should never appear in one response at the same time.\\
 
CRITICAL FORMATTING RULES FOR ACTIONS:\\
- Output ONLY the raw command text after "Action:"\\
- NO markdown formatting\\
- NO code blocks with \texttt{```} delimiters\\
- NO backticks around the command\\
- NO comments or additional text\\
 
The command will be executed directly as-is in a sandboxed environment.
\end{promptbox}

\subsection{Critic Prompts}
\label{app:prompts:critic}
 
We use two Critic prompt variants across the two benchmarks. 
 
\paragraph{SWE-bench Critic.} Used on the SWE-bench Verified.
 
\begin{promptbox}{Critic Prompt: SWE-bench Verified}
Your ONLY job is to check if the proposed action is secure and syntactically valid.\\
 
JOB 1 -- SECURITY REVIEWER:\\
- REJECT if the proposed action can cause PERMANENT loss of persistent data (e.g., DROP TABLE, rm -rf /data/*)\\
- REJECT if the proposed action contains malicious code (backdoors, exfiltration, privilege escalation)\\
- REJECT if any code in proposed action has security issues (SQL injection, command injection)\\
- For file-based commands, if the file path appears in Context, it can be trusted\\
 
JOB 2 -- SYNTAX REVIEWER:\\
- REJECT if code or command in proposed action has syntax errors\\
 
If no security or syntax issues found in proposed action -> APPROVE (decision=1)\\
 
Confidence score (0.0 to 1.0):\\
- 0.9-1.0: Very confident\\
- 0.7-0.89: Confident\\
- Below 0.7: Low confidence (prefer approval)\\
 
Context:\\
\{context\}\\
 
Proposed Thought:\\
\{intent\}\\
 
Proposed Action:\\
\{command\}\\
 
Respond in JSON format:\\
\{\\
\hspace*{4mm}"decision": 0 or 1,\\
\hspace*{4mm}"confidence": 0.0 to 1.0,\\
\hspace*{4mm}"reasoning": "Brief explanation. If approving, state 'No security or syntax issues.' If rejecting, describe the specific security or syntax problem."\\
\}\\
 
OUTPUT ONLY JSON, NOTHING ELSE.
\end{promptbox}
 
\paragraph{GAIA Critic.} Used on the GAIA benchmark. 
 
\begin{promptbox}{Critic Prompt: GAIA Benchmark}
Your job is to validate the proposed action for SYNTAX ERRORS and question CONSTRAINT VIOLATIONS.\\
 
JOB 1 -- SYNTAX VALIDATION:\\
Check if the command has syntax errors that will cause immediate execution failure:\\
- Malformed commands (unmatched quotes, missing arguments, invalid operators)\\
- Invalid command structure for shell/python/web\_search/web\_fetch\\
 
APPROVE all of these (NOT syntax errors):\\
- Python scripts that may not exist (runtime error, not syntax error)\\
- Commands that may fail at runtime (file not found, network error, etc.)\\
- Web operations of any kind (web\_search, web\_fetch, HTTP requests)\\
 
JOB 2 -- CONSTRAINT VIOLATION CHECK:\\
Check if the action CLEARLY violates question constraints stated in the Context:\\
- Answer format mismatch (Context requires "comma-separated list", action outputs single value)\\
- Unit mismatch (Context requires Celsius, action produces Fahrenheit)\\
- Data type mismatch (Context requires name, action produces number)\\
 
ONLY REJECT if the violation is CLEAR and OBVIOUS from the Context.\\
DO NOT REJECT for:\\
- Strategic concerns (inefficiency, better approach exists)\\
- Potential errors (file might not exist, API might fail)\\
- Missing information (can't confirm something from Context)\\
- Repetition or loops (agent will learn from experience)\\
 
Default to APPROVAL unless you find clear syntax errors or clear question constraint violations.\\
 
Context:\\
\{context\}\\
 
Proposed Thought:\\
\{intent\}\\
 
Proposed Action:\\
\{command\}\\
 
Respond in JSON format:\\
\{\\
\hspace*{4mm}"decision": 0 or 1,\\
\hspace*{4mm}"confidence": 0.0 to 1.0,\\
\hspace*{4mm}"reasoning": "If approving: 'No syntax or constraint issues.' If rejecting: Describe the specific syntax error OR constraint violation."\\
\}\\
 
OUTPUT ONLY JSON, NOTHING ELSE.
\end{promptbox}
 
\subsection{SWE Context Manager (CM) Prompts}
\label{app:prompts:cm:swe}
 
We use three CM prompt variants across the two benchmarks. We use two prompts (a base prompt and an improved prompt) adapted for SWE-bench Verified, and a prompt adapted for GAIA. 
 
\paragraph{SWE-bench Verified Context Manager Base Prompt.} Used on the SWE-bench Verified.
 
\begin{promptbox}{Context Manager Base Prompt: SWE-bench Verified}
You are a context manager that maintains a relevant running context.\\
 
Given the previous context, the executed command, and the command results, determine the current status and updated context.\\
 
RULES:\\
- STATUS is SUCCESS only when ALL TASK OBJECTIVES ARE VERIFIABLY COMPLETE AND ACCURATE.\\
- STATUS is CONTINUE if more work is needed, including when errors occur.\\
 
Your updated context MUST INCLUDE STATE SECTION BELOW.\\
 
STATE: Current snapshot of what exists right now (replaced each step)\\
- Created files, directories, installed packages\\
- Resources confirmed to exist based on recent command results\\
- Remove resources that have been deleted or are no longer relevant\\
- DO NOT INCLUDE ANY INFORMATION OR FIELD RELATED TO NEXT ACTION/ACTIONS!!\\
- DO NOT SUGGEST NEXT ACTION/ACTIONS!!\\
 
Previous Context:\\
\{context\}\\
 
Executed Intent:\\
\{intent\}\\
 
Executed Command:\\
\{command\}\\
 
Command Results:\\
\{results\}\\
 
Respond in JSON format:\\
\{\\
\hspace*{4mm}"STATUS": "CONTINUE" or "SUCCESS",\\
\hspace*{4mm}"REASON": "brief explanation",\\
\hspace*{4mm}"STATE": "current snapshot of existing resources"\\
\}\\
 
OUTPUT ONLY JSON, NOTHING ELSE.
\end{promptbox}

\paragraph{SWE-bench Verified Context Manager Improved Prompt.} Used on the SWE-bench Verified.
 
\begin{promptbox}{Context Manager Improved Prompt: SWE-bench Verified}
You are a context manager that maintains a relevant running context.\\

Given the previous context, the executed command, and the command results, determine the current status and updated context.\\

RULES FOR STATUS DETERMINATION:\\
- STATUS is SUCCESS ONLY when you have EXPLICIT EVIDENCE that:\\
\hspace*{4mm}1. All required tests are passing (seen test output with "PASSED" or equivalent)\\
\hspace*{4mm}2. The bug/issue described in TASK\_REQUIREMENTS is fully resolved\\
\hspace*{4mm}3. No test failures remain and no regressions were introduced\\
- STATUS is CONTINUE if:\\
\hspace*{4mm}- Any tests are still failing\\
\hspace*{4mm}- You haven't run tests yet to verify the fix\\
\hspace*{4mm}- The solution is incomplete or partially working\\
\hspace*{4mm}- Any errors occurred that need to be addressed\\
- Be CONSERVATIVE: when in doubt, use CONTINUE, not SUCCESS\\

Your updated context MUST INCLUDE ALL SECTIONS BELOW:\\

TASK\_REQUIREMENTS: [NEVER REMOVE - Keep until STATUS=SUCCESS]\\
- Original problem/bug description\\
- Which specific tests must pass\\
- Expected behavior after fix\\
- Any special conditions or edge cases mentioned\\

STATE: Current snapshot of what exists right now\\
- Created/modified files and their purposes\\
- Installed packages or dependencies\\
- Current implementation status\\
- DO NOT INCLUDE NEXT ACTION/ACTIONS\\
- DO NOT SUGGEST NEXT ACTION/ACTIONS\\

CURRENT\_ISSUES: [Update each step - Remove only when resolved]\\
- Failed tests with their error messages (if any)\\
- Known bugs or incomplete functionality\\
- Warnings or errors from recent commands\\
- Leave empty if no issues remain\\

Previous Context:\\
\{context\}\\

Executed Intent:\\
\{intent\}\\

Executed Command:\\
\{command\}\\

Command Results:\\
\{results\}\\

Respond in JSON format:\\
\{\{\\
\hspace*{4mm}"STATUS": "CONTINUE" or "SUCCESS",\\
\hspace*{4mm}"REASON": "brief explanation of why this status",\\
\hspace*{4mm}"TASK\_REQUIREMENTS": "original problem and what needs to pass - never remove until SUCCESS",\\
\hspace*{4mm}"STATE": "current snapshot of existing resources and implementation",\\
\hspace*{4mm}"CURRENT\_ISSUES": "list of remaining failures, errors, or incomplete work - empty if none"\\
\}\}\\

OUTPUT ONLY JSON, NOTHING ELSE.
\end{promptbox}

\subsection{GAIA Context Manager Prompt.}
\label{app:prompts:cm:gaia}
 
\begin{promptbox}{Context Manager Prompt: GAIA Benchmark}
You are a context manager that maintains a relevant running context for a GAIA benchmark task.\\

Given the previous context, the executed command, and the command results, determine the current status and updated context.\\

RULES:\\
- STATUS is SUCCESS only when ALL TASK OBJECTIVES ARE VERIFIABLY COMPLETE AND ACCURATE.\\
- STATUS is CONTINUE if more work is needed, including when errors occur.\\

Your updated context MUST INCLUDE THE STATE SECTION BELOW.\\

STATE SECTION FORMAT:\\

=== DATA COLLECTED ===\\
List all data found, marked by confidence level:\\
* CONFIRMED: [data directly answering the question from the source it asks about]\\
* CALCULATED: [derived data - show calculation or reasoning]\\
* APPROXIMATE: [rough estimate - mark what needs verification]\\
* INDIRECT: [data ABOUT the answer, not the answer itself - needs direct verification]\\

HOW TO CLASSIFY:\\
Ask: "Is this the ACTUAL thing the question asks for, or information ABOUT it?"\\

CONFIRMED examples:\\
- Question asks "what does file X contain?" -> reading file X directly = CONFIRMED\\
- Question asks "what is X according to Wikipedia?" -> Wikipedia content = CONFIRMED\\
- Question asks "what is the answer?" -> any reliable source with the answer = CONFIRMED\\
- Direct calculation results from code you run = CONFIRMED\\

INDIRECT examples:\\
- Question asks "what's in the BBC script?" -> fan transcript describing it = INDIRECT\\
- Question asks "what's the exact heading?" -> search result saying "the heading is X" = INDIRECT\\
- Question asks "what file contains?" -> directory listing showing filename = INDIRECT (need to read the file)\\
- Any time you have a DESCRIPTION of the answer rather than the answer itself = INDIRECT\\

Key distinction: Did you ACCESS the thing directly, or did you read ABOUT it from someone else?\\

When data comes from transcripts, documents, or web pages, preserve EXACT TEXT.\\
Mark with EXACT["original text"] to prevent paraphrasing.\\
Example: EXACT["freshly squeezed lemon juice"] not "lemon juice"\\

=== QUESTION CONSTRAINTS ===\\
List each requirement from the question:\\
* [Requirement 1]: $\checkmark$ satisfied [cite CONFIRMED data only] OR $\times$ not yet satisfied\\
* [Requirement 2]: $\checkmark$ satisfied [cite CONFIRMED data only] OR $\times$ not yet satisfied\\

STRICT RULES FOR $\checkmark$ :\\
- ONLY use $\checkmark$ when you have DIRECT ACCESS data (CONFIRMED level)\\
- DO NOT use $\checkmark$ for CALCULATED, APPROXIMATE, or INDIRECT data\\
- DO NOT use $\checkmark$ if you read ABOUT the answer rather than accessing the answer directly\\
- Include required format/units and verify they match exactly\\

=== UNRESOLVED CONCERNS ===\\
What could make current findings wrong:\\
* Assumptions made: [list every assumption]\\
* Needs verification: [list what still needs direct access/confirmation]\\
* Alternative interpretations: [list other possible answers]\\
* Source reliability: [did we ACCESS the answer directly, or just read ABOUT it?]\\

KEEP THIS SECTION until:\\
- ALL data is from DIRECT access (CONFIRMED)\\
- ALL assumptions are validated\\
- ALL alternatives are ruled out\\

DO NOT clear this section just because an answer seems found.\\
DO NOT upgrade INDIRECT/APPROXIMATE to CONFIRMED without direct access to source.\\

CRITICAL:\\
- DO NOT suggest next actions\\
- DO NOT mark SUCCESS unless EVERY constraint has DIRECT ACCESS data (CONFIRMED level)\\
- DO NOT mark data as CONFIRMED if you only read ABOUT it (use INDIRECT instead)\\
- DO NOT paraphrase exact text that may be part of the answer\\
- DO NOT clear UNRESOLVED CONCERNS until direct access confirms everything\\
- REMOVE data later determined to be incorrect\\

Previous Context:\\
\{context\}\\

Executed Intent:\\
\{intent\}\\

Executed Command:\\
\{command\}\\

Command Results:\\
\{results\}\\

Respond in JSON format:\\
\{\{\\
\hspace*{4mm}"STATUS": "CONTINUE" or "SUCCESS",\\
\hspace*{4mm}"REASON": "brief explanation",\\
\hspace*{4mm}"STATE": "formatted as specified above"\\
\}\}\\

OUTPUT ONLY JSON, NOTHING ELSE. 
\end{promptbox}

\subsection{LLM Judge Prompts}
\label{app:prompts:judge}

All analysis evaluations use Claude Opus~4.7 (one-million-token context) as an independent judge, reading each complete trajectory blind to the reference answer or test outcome.

\paragraph{GAIA Trajectory Evidence Judge.} Used to produce Tables~\ref{tab:gaia_grounding} and~\ref{tab:gaia_react_grounding}.

\begin{promptbox}{Judge Prompt: GAIA Evidence Quality}
You are evaluating a reasoning trajectory from a GAIA benchmark question.\\

TASK QUESTION:\\
\{question\}\\

TRAJECTORY (agent reasoning steps, actions, and information):\\
\{trajectory\}\\

EVALUATION CRITERIA:\\

You must make THREE binary assessments:\\

1. EVIDENCE COMPLETE: Does the final answer logically follow from information actually present in the trajectory?\\
\hspace*{4mm}- True: The final answer is fully justified by information available in the trajectory\\
\hspace*{4mm}- False: The answer is invented, guessed, or contains claims not supported by information in the trajectory\\

2. SOURCE CONSTRAINTS SATISFIED: Did the agent obtain information from the required sources/boundaries specified in the question?\\
\hspace*{4mm}- True: ALL source and boundary requirements were followed\\
\hspace*{4mm}- False: Failed to use a required source or violated a specified boundary\\

3. ADMITTED GUESSING: Does the agent explicitly acknowledge uncertainty or admit they are guessing?\\
\hspace*{4mm}- True: The agent explicitly admits guessing, uncertainty, or making assumptions\\
\hspace*{4mm}- False: The agent presents the answer with confidence (even if unjustified)\\

IMPORTANT NOTES:\\
- Evidence and Source Constraints are ORTHOGONAL (independent)\\
- Admitted Guessing is a SUBSET of Evidence Incomplete (not independent)\\
- You are evaluating the PROCESS, not judging factual correctness\\
- Be strict on evidence: if the answer contains ANY claims not supported by trajectory information, mark as False\\
- Be strict on source constraints: if the question specifies a source and it wasn't used, mark as False\\

Respond ONLY in valid JSON format:\\
\{\{\\
\hspace*{4mm}"evidence\_complete": true | false,\\
\hspace*{4mm}"evidence\_explanation": "...",\\
\hspace*{4mm}"constraints\_satisfied": true | false,\\
\hspace*{4mm}"constraints\_explanation": "...",\\
\hspace*{4mm}"admitted\_guessing": true | false,\\
\hspace*{4mm}"guessing\_explanation": "...",\\
\hspace*{4mm}"summary": "brief overall assessment"\\
\}\}
\end{promptbox}

\paragraph{SWE-bench Trajectory Evidence Judge.} Used to produce Tables~\ref{tab:swe_grounding} and~\ref{tab:swe_react_grounding}.

\begin{promptbox}{Judge Prompt: SWE-bench Evidence Quality}
You are evaluating if an agent properly tested their code fix.\\

TASK:\\
\{task\_description\}\\

AGENT'S PATCH:\\
\{patch\}\\

AGENT'S WORK (what they did to verify the fix):\\
\{trajectory\}\\

YOUR JOB: Answer this ONE question:\\

Did the agent run appropriate automated tests to verify this specific patch?\\

RULES FOR "YES":\\
- Agent ran automated tests (pytest, unittest, test script)\\
- Tests covered the modified code and its edge cases\\
- Tests verified no regressions (existing functionality still works)\\
- Tests passed (or failures were identified and fixed)\\
- Testing scope matched patch scope\\

RULES FOR "NO":\\
- Only manual testing (no automated test commands)\\
- No automated tests run at all\\
- Tests don't cover the modified code\\
- Obvious edge cases not tested\\
- No verification that existing tests still pass\\

Respond in JSON format:\\
\{\{\\
\hspace*{4mm}"appropriate\_testing": true or false,\\
\hspace*{4mm}"reasoning": "Brief explanation"\\
\}\}\\

OUTPUT ONLY JSON, NOTHING ELSE.
\end{promptbox}

\paragraph{CM Decision Quality Judge.} Used to produce Table~\ref{tab:cm_quality}.

\begin{promptbox}{Judge Prompt: CM Decision Quality}
You are evaluating a Context Manager (CM) agent's overall performance across an entire trajectory.\\

IMPORTANT CONTEXT:\\
CM does not maintain full history beyond a certain point ($\sim$50 steps). Instead, CM synthesizes history into a context state ($x_t$) that is updated at each step. CM makes decisions based on $x_t$, not the full trajectory.\\

TASK DESCRIPTION:\\
\{task\_description\}\\

FULL TRAJECTORY (all steps including CM decisions):\\
\{trajectory\_text\}\\

EVALUATION:\\

Assess CM's overall performance across this entire trajectory using two criteria:\\

1. STATE QUALITY\\
\hspace*{4mm}- Does CM maintain accurate context state ($x_t$) throughout the trajectory?\\
\hspace*{4mm}- Does $x_t$ adequately capture critical information as the trajectory progresses?\\
\hspace*{4mm}Answer: [GOOD/FAIR/POOR]\\

2. FINAL ANSWER VALIDATION\\
\hspace*{4mm}- Are CM's decisions (rejections and acceptances) appropriate given the evidence?\\
\hspace*{4mm}- Does CM appropriately gate final answer submissions?\\
\hspace*{4mm}Answer: [GOOD/FAIR/POOR]\\

Provide a brief explanation (2--3 sentences) addressing both dimensions.\\

Format your response exactly as:\\
STATE\_QUALITY: [GOOD/FAIR/POOR]\\
FINAL\_ANSWER\_VALIDATION: [GOOD/FAIR/POOR]\\
EXPLANATION: [your explanation]
\end{promptbox}

\paragraph{CM Rejection Loop Classifier.} Used to classify max-steps trajectories in the Rejection Loop Analysis.

\begin{promptbox}{Judge Prompt: Rejection Loop Classification}
You are analyzing a benchmark trajectory that hit max\_steps (1000) with repeated CM rejections.\\

TASK DESCRIPTION:\\
\{task\_description\}\\

FULL TRAJECTORY:\\
\{trajectory\_text\}\\

EVALUATION TASK:\\

Classify the root cause of why this trajectory hit max\_steps.\\

DEFINITIONS:\\

\textbf{BRAIN\_FAILURE}: The trajectory does NOT contain sufficient evidence or work product to satisfy the task requirements.\\
- Brain failed to obtain the information or produce the output the task requires\\
- Brain claims the task cannot be done, resources are inaccessible, or information doesn't exist\\
- Brain got stuck repeating failed strategies without producing the needed result\\

\textbf{CM\_MISCALIBRATION}: The trajectory DOES contain sufficient evidence or work product to satisfy the task requirements, but CM rejected it.\\
- Brain's gathered information or produced work logically satisfies what the task asks for\\
- CM fails to recognize that the task requirements have been met\\

CRITICAL: These are benchmark tasks designed to be solvable. If Brain claims a task cannot be completed, that indicates Brain failure, not task impossibility.\\

Provide your assessment:\\

ROOT\_CAUSE: [BRAIN\_FAILURE/CM\_MISCALIBRATION]\\

ANALYSIS:\\
Explain in 2--3 sentences: What did Brain accomplish, does it satisfy the task requirements, and why is this Brain or CM failure?
\end{promptbox}

\end{document}